\documentclass[twoside]{article}

\usepackage[accepted]{aistats2023}
\usepackage[round]{natbib}

\usepackage[hidelinks]{hyperref}
\usepackage{graphicx}
\newcommand*\rot{\rotatebox{90}}
\usepackage{multirow}
\usepackage{xcolor}

\usepackage{pifont}
\usepackage{amsmath}
\usepackage{amsfonts}
\usepackage{float}
\usepackage{enumitem}
\usepackage[stable]{footmisc}

\usepackage{caption} 
\usepackage{longtable}

\begin{document}

%

%

\twocolumn[

\aistatstitle{Vector Quantized Time Series Generation with a Bidirectional Prior Model}

\aistatsauthor{Daesoo Lee \And Sara Malacarne \And  Erlend Aune}

\aistatsaddress{Norwegian University-\\of Science and Technology \And  Telenor Research \And Norwegian University-\\of Science and Technology \\ BI Norwegian Business School \\ Abelee} ]

\begin{abstract}
Time series generation (TSG) studies have mainly focused on the use of Generative Adversarial Networks (GANs) combined with recurrent neural network (RNN) variants. However, the fundamental limitations and challenges of training GANs still remain. In addition, the RNN-family typically has difficulties with temporal consistency between distant timesteps. Motivated by the successes in the image generation (IMG) domain, we propose TimeVQVAE, the first work, to our knowledge, that uses vector quantization (VQ) techniques to address the TSG problem. Moreover, the priors of the discrete latent spaces are learned with bidirectional transformer models that can better capture global temporal consistency. We also propose VQ modeling in a time-frequency domain, separated into low-frequency (LF) and high-frequency (HF). This allows us to retain important characteristics of the time series and, in turn, generate new synthetic signals that are of better quality, with sharper changes in modularity, than its competing TSG methods. Our experimental evaluation is conducted on all datasets from the UCR archive, using well-established metrics in the IMG literature, such as Fréchet inception distance and inception scores. 
Our implementation on GitHub: \url{https://github.com/ML4ITS/TimeVQVAE}.
\end{abstract}

\section{INTRODUCTION}
\label{sect:introduction}
TSG models have been developed to overcome insufficient training data due, for example, to acquisition difficulties or strict privacy constraints \citep{li2015generative, esteban2017real, yoon2019time, ni2020conditional, smith2020conditional, li2022tts, zha2022time}. Current mainstream TSG methods consist of combining RNNs with the GAN architecture, as done in RCGAN \citep{esteban2017real}, TimeGAN \citep{yoon2019time}, and SigCWGAN \citep{ni2020conditional}. However, such methods are unable to effectively produce long sequences, given the limitation of the RNN model in keeping track of temporal dependencies between inputs that are distant in time \citep{vaswani2017attention}. This problem is tackled by \cite{li2022tts} with TTS-CGAN (Transformer Time-Series Conditional GAN), where the RNN model is replaced by a transformer model \citep{vaswani2017attention}. The challenges of training GANs, such as non-convergence, mode collapse, generator-discriminator unbalance, and sensitive hyperparameter selection, however, still remain.

There has been much evidence in the IMG literature that VQ-based models and diffusion models significantly outperform the GAN alternatives \citep{yu2022scaling, gafni2022make, chang2022maskgit, saharia2022photorealistic, ramesh2022hierarchical}. Therefore, a meaningful performance gain is to be expected by extending analogous methodologies in the time series domain.  With our proposal, which we call TimeVQVAE, we adopt the two-stage modeling approach appearing in \citep{van2017neural}. More specifically, VQ-VAE is used for the first stage and MaskGIT \citep{chang2022maskgit} for the second. Although there are other advanced approaches for the first stage, such as VQ-GAN \citep{esser2021taming} and ViT-VQGAN \citep{yu2021vector}, VQ-VAE has been selected as it has shown good performance in several areas, and, to our knowledge, it has never been explored in the TSG domain. The advanced VQ methods could be further investigated in future work. MaskGIT proposes to use a bidirectional transformer model for the prior learning unlike VQ-VAE, VQ-GAN, and ViT-VQGAN, where autoregressive models are used. It has experimentally been shown that the bidirectional transformer can capture global consistency better and accelerates the data generation process and achieves better-quality synthetic samples. 

In our work, VQ-VAE acts on the time-frequency domain. We, furthermore, propose to separate the VQ modeling into LF and HF, respectively. The time-frequency domain provides richer information, results in better VQ modeling than in the time domain, and it allows disentangling HF components, such as spikes, from LF components, such as trends, more easily. 
Frequency separation is also motivated by the different predictability and compressibility of LF and HF components, as LF components are more predictable and compressible than HF components by nature. Thus, we let our model generate the LF component first and only subsequently the HF component, which could be viewed as defining the overall shape first and then filling in the details.
Frequency separation also allows us to more precisely address several useful downstream tasks, such as anomaly detection for LF and HF components respectively, with explainable restoration \citep{marimont2021anomaly}. Another related task is time series forecasting using discrete latent vectors, as explored in \citep{rasul2022vq}, where an autoregressive model is trained in the discrete latent space, showing competitive forecasting performance. With the proposed LF-HF separation, the VQ forecasting problem can be further eased: LF components can be easily predicted while HF components can be quantified for uncertainty using the likelihood of individual tokens.
In our proposed method, TimeVQVAE, we use STFT to transform the time series into the time-frequency domain and split it into LF and HF regions. Then, two sets of encoder, decoder, and vector-quantizer are used to learn the discrete latent spaces for LF and HF, respectively. Next, priors of the LF and HF discrete latent spaces are learned with two bidirectional transformer models. Lastly, we propose to jointly sample sets of LF and HF discrete latent vectors from the learned priors and decode them into time series with the learned decoders. 

To evaluate generated samples fairly, robust evaluation metrics and diverse benchmark datasets are necessary. There has, however, been a lack of agreement for which evaluation metrics to use in the TSG literature \citep{brophy2021generative} and most TSG studies have been evaluated only on a small range of datasets \citep{esteban2017real, yoon2019time, ni2020conditional, li2022tts, zha2022time, desai2021timevae}.
A common visual comparative evaluation of TSG is carried out using PCA and t-SNE on time series \citep{yoon2019time, desai2021timevae, zha2022time, li2022tts}. But because this cannot be reported as a single value metric, its usability is limited. 
Also, because each element of time series does not carry semantically-meaningful information but time-step information, the principal axes found by PCA are not effective in capturing the realism of time series.
There are other metrics that are occasionally used, such as the discriminative and predictive scores \citep{yoon2019time, desai2021timevae, zha2022time}, but there is no consensus for those yet, as they lack stability in the sense that the scores can be largely inconsistent for the same methods across different studies \citep{yoon2019time, desai2021timevae, zha2022time}. 
The IMG literature has already established standard metrics, such as the inception score (IS) \citep{salimans2016improved}, Fréchet inception distance (FID) score \citep{heusel2017gans}, and Classification Accuracy Score (CAS) \citep{ravuri2019classification}. Note that CAS is the same as TSTR (Training on Synthetic data and Testing on Real data). 

In this work, we propose to use the above-listed metrics -- IS and FID in particular -- to evaluate TSG models on a wide range of diverse datasets from the UCR archive \citep{UCRArchive2018} as an evaluation protocol for TSG models. 
IS and FID score have rarely been used in the TSG literature because there is no pretrained model for time series unlike in the computer vision domain \citep{NEURIPS2019_9015}. 
\cite{smith2020conditional} was the first to evaluate a TSG model in terms of FID scores over the UCR archive datasets. They trained the Fully Convolutional Network (FCN) model \citep{wang2017time} on every available UCR dataset and computed FID scores using the pretrained FCN models. Because the FCN model is one of the strongest baselines for time series classification and can effectively capture pattern features of time series \citep{wang2017time}, it is reasonable to use the representations from such a model to compute the FID scores. Unfortunately, though, the pretrained FCN models are not provided by \cite{smith2020conditional} on any open-source platform. We thus follow the same protocol as \citep{smith2020conditional} and contribute by sharing code for the pretrained FCN models on GitHub along with templates for computing the IS and FID scores; available on \url{https://github.com/danelee2601/supervised-FCN}. It can be easily installed via pip. 

Experiments are conducted for unconditional sampling and class-conditional sampling over the entire UCR archive. Our results clearly demonstrate how our proposed method outperforms the current competing TSG models, such as GMMN \citep{li2015generative}, RCGAN, TimeGAN, SigCWGAN, and TSGAN, in terms of IS, FID, and CAS.

To summarize, our contributions consist of:
\begin{enumerate}[topsep=0pt,itemsep=-1ex,partopsep=1ex,parsep=1ex]
    \item[$\bullet$] use of VQ for TSG,
    \item[$\bullet$] VQ modeling in the time-frequency domain,
    \item[$\bullet$] latent space separation into LF and HF,
    \item[$\bullet$] joint sampling from the LF and HF latent spaces,
    \item[$\bullet$] guided class-conditional sampling,
    \item[$\bullet$] TSG evaluation on the entire UCR archive, 
    \item[$\bullet$] evaluation conducted by reporting IS, FID, and CAS,
    \item[$\bullet$] releasing the pretrained FCN models on the UCR archive.
\end{enumerate}

\section{RELATED WORK}

\subsection{Time Series Generation}
\paragraph{GMMN} employs maximum mean discrepancy (MMD). MMD is utilized to partially overcome the unstable training problem of GAN. GMMN has a simple objective that can be interpreted as matching all orders of statistics between real samples and synthetic samples. For TSG, GMMN can be combined with an RNN model. The MMD loss can be minimized between the generated time series by the RNN model and the real time series \citep{sigcwgan_github}.

\paragraph{RCGAN} sequentially generates time series using GAN: a latent vector from the noise space is sampled at every temporal step and an RNN model sequentially encodes them generating a synthetic time series. 

\paragraph{TimeGAN} uses the conventional GAN training combined with a supervised learning approach. It aims to better preserve temporal dynamics of time series with the supervised learning loss. 

\paragraph{SigCWGAN} addresses the problem of GANs struggling with capturing temporal dependencies by using the signature of a path \citep{kidger2019deep}. \cite{ni2020conditional} proposes a Signature Wasserstein-1 (Sig-$W_1$) metric that better captures temporal dependencies, and uses it as a discriminator. It provides a universal description of complex data streams without requiring expensive computation like the Wasserstein metric.

\paragraph{TTS-CGAN} uses a transformer model to overcome the distant temporal dependency problem appearing for the TSG RNN-based GAN models, such as RCGAN, TimeGAN, and SigCWGAN. It also proposes an approach to better induce the class-conditioning information into generated samples. Yet, the fundamental challenges in adopting GANs -- that is, non-convergence, mode collapse, unbalance between generator and discriminator, sensitive hyperparameter selection -- still exist \citep{arjovsky2017towards, salimans2016improved, goodfellow2016nips, goodfellow2020generative, lucic2018gans}.

\paragraph{TSGAN} is built using two WGANs \citep{arjovsky2017wasserstein}. The first WGAN generates a synthetic spectrogram from the noise latent vector, and the second WGAN receives the synthetic spectrogram and generates a time series. Therefore, there exist two discriminators. The first discriminator compares the spectrograms while the second discriminator compares the time series. Unlike RCGAN, TimeGAN, and SigCWGAN which use RNN-based generators, TSGAN's generators use convolutional layers.

\subsection{Vector Quantization-based Image Generation}

    

\paragraph{VQ-VAE} is a type of variational autoencoder that uses vector quantization to obtain a discrete latent representation. There are two main differences between the classical VAE and the VQ-VAE: 1) VQ-VAE maps an input to a discrete latent space rather than a continuous space, and 2) instead of being static, the prior is learned. Because there is no constraint on the prior being Gaussian, as in VAE, VQ-VAE does not suffer from the posterior collapse. Moreover, the discretization allows for sharp and crisp synthetic images. For the sampling process, \cite{van2017neural} proposes the two-stage modeling approach: The first stage is for learning to project an input to the discrete latent space, that is, the so-called tokenization. To do that, an encoder, decoder, and codebook are trained. The codebook stores $K$ discrete codes (= discrete latent vectors/tokens). The second stage is for learning a prior of the discrete tokens in the discrete latent space. In VQ-VAE, PixelCNN \citep{van2016pixel} and WaveNet \citep{oord2016wavenet} are used as prior models for images and audio data, respectively.

\paragraph{MaskGIT} proposes a bidirectional transformer model for the prior learning instead of an autoregressive model as in VQ-VAE, VQ-GAN, and ViT-VQGAN. MaskGIT trains the prior model in the masked modeling manner. \cite{chang2022maskgit} shows that using the bidirectional transformer model can accelerate the sampling process by 30-64 times and can achieve better quality and more diversity in the synthetic samples.

\paragraph{VQ-based Text-to-Image Generative Models} The current state-of-the-art (SOTA) IMG models include VQ and diffusion models, and not GANs. Among the SOTA VQ-based IMG models there are Parti \citep{yu2022scaling}, Make-A-Scene \citep{gafni2022make}, and DALLE-E \citep{ramesh2021zero}. Their performances are quite competitive to their diffusion-based counterparts, such as Imagen \citep{saharia2022photorealistic}, DALL-E-2 \citep{ramesh2022hierarchical}, and GLIDE \citep{nichol2021glide}. In light of such improvements in the image domain, a performance gain is to be expected also in the time series domain, if adopting one of the above-mentioned methodologies. With such motivation, our work explores the VQ-VAE approach for addressing the TSG problem.

\section{METHOD}
To produce good quality synthetic samples, optimization in stage~1 (tokenization) and stage~2 (prior learning) needs to be ensured. In the following subsections, our proposals for stages 1 and 2 are presented.


\begin{figure*}[]
\centering
\includegraphics[width=0.99\linewidth]{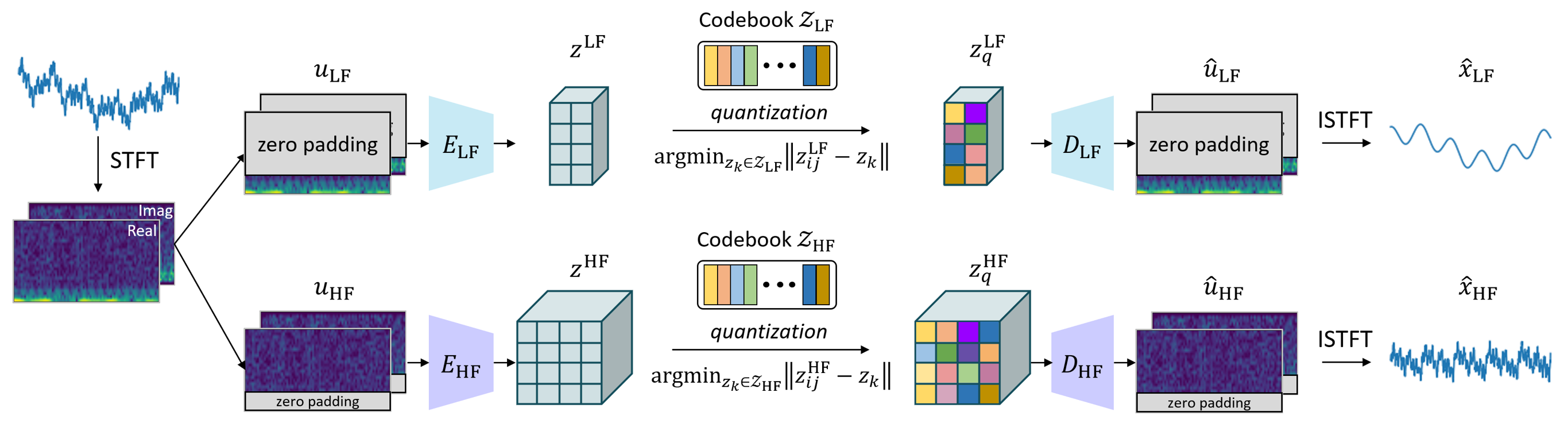}
\caption{Overview of our proposed VQ (\textit{i.e.,} tokenization) (stage 1).}
\label{fig:overview_timevqvae}
\end{figure*}

\subsection{Stage 1: Learning Vector Quantization}
In stage 1, an encoder, decoder, and codebook are to be optimized by effectively compressing the input into discrete tokens, while minimizing the information loss obtained by comparing the input with the output, given by decoding the selected tokens.
In our work, VQ-VAE is used as a basis. VQ-VAE is known to produce sharper reconstructions \citep{van2017neural} than AE and VAE \citep{bank2020autoencoders, podgorskiy_VAE}. However, we have experimentally found that a naive form of VQ-VAE still experiences difficulty with reconstructing time series, especially HF components. We tackle this problem by 1) augmenting the time series into the time-frequency domain, 2) separating the latent space modeling into LF and HF. By doing so, we can compress both LF and HF information of time series into the latent space better.

The overview of our proposal for stage~1 is presented in Fig.~\ref{fig:overview_timevqvae}. The encoder and the decoder are denoted by $E$ and $D$ respectively. STFT and ISTFT stand for Short-time Fourier Transform and Inverse Short-time Fourier Transform, respectively. First, a time series is \textit{augmented} into the time-frequency domain and separated into two branches where one is zero-padded on the HF region and the other is zero-padded on the LF region. Then, the encoders -- $E_{\mathrm{LF}}$ and $E_{\mathrm{HF}}$ -- project the time-frequency domains into the continuous latent space. 
$E_{\mathrm{LF}}$ has a higher downsampling rate than $E_{\mathrm{HF}}$. The higher downsampling rate has a larger receptive field, therefore, enables to capture the overall structure of the data better, which results in globally-consistent synthetic samples \citep{esser2021taming}. But such a high downsampling rate fails to retain the HF information in the embedding space \citep{rombach2022high}. To produce globally-consistent synthetic data with fine HF details, we use $E_{\mathrm{LF}}$ with a higher downsampling rate and $E_{\mathrm{HF}}$ with a lower downsampling rate.
The continuous latent space is further transformed into the discrete latent space by the codebook via the argmin process. In the argmin process, each continuous token is compared to every discrete token in the codebook in terms of the Euclidean distance and replaced with the closest discrete token. That is the so-called \textit{quantization}. Then, the decoders project the discrete latent spaces back into the time-frequency domains equipped with the corresponding zero-paddings, which are then mapped to the time domains via ISTFT. At the end, the two branches produce LF and HF components of the time series, respectively -- $\hat{x}_{\mathrm{LF}}$ and $\hat{x}_{\mathrm{HF}}$. 

The codebook $\mathcal{Z}$ consists of $K$ discrete tokens $\mathcal{Z} = \{z_k\}_{k=1}^{K}$, where each $z_k \in \mathbb{R}^{d}$ and $d$ denotes dimension size. The quantization process can be formulated as 
\begin{equation}
\label{eq:z_q}
(z_q)_{ij} = \operatorname*{argmin}_{z_k \in \mathcal{Z}} \|z_{ij} - z_k\|,
\end{equation}
where $z \in \mathbb{R}^{d \times h \times w}$ denotes the activation map after the encoder, in which $h$ and $w$ represent the height and width of the map, respectively. 
For each element in $z$, denoted by $i,j$, the corresponding continuous tokens are denoted by $z_{ij}\in \mathbb{R}^{d}$.
The codebook-learning loss is then given by
\begin{equation}
\label{eq:codebook_learning_loss}
\begin{split}
\mathcal{L}_{\mathrm{codebook}} & = \| \mathrm{sg}[E_{\mathrm{LF}}( \mathcal{P}_{\mathrm{LF}}(\mathrm{STFT}(x)) )] - z_q^{\mathrm{LF}} \|_2^2 \\
                                & + \| \mathrm{sg}[E_{\mathrm{HF}}( \mathcal{P}_{\mathrm{HF}}(\mathrm{STFT}(x)) )] - z_q^{\mathrm{HF}} \|_2^2 \\
                                & + \beta \| E_{\mathrm{LF}}( \mathcal{P}_{\mathrm{LF}}(\mathrm{STFT}(x)) ) - \mathrm{sg}[z_q^{\mathrm{LF}}] \|_2^2 \\
                                & + \beta \| E_{\mathrm{HF}}( \mathcal{P}_{\mathrm{HF}}(\mathrm{STFT}(x)) ) - \mathrm{sg}[z_q^{\mathrm{HF}}] \|_2^2,
\end{split}
\end{equation}
where $x$ denotes the time series, $\mathrm{sg}[\cdot]$ denotes the stop-gradient operation, $\mathcal{P}_{\mathrm{[\cdot]}}$ denotes the zero-padding operation on either the LF or HF region, $z_q^{[\cdot]}$ denotes the discrete token for either LF or HF, and $\beta$ is a weighting parameter for the commitment loss terms.
The back-propagation through the non-differentiable quantization is achieved simply by copying the gradients from the decoder to the encoder.

The VQ loss also contains the reconstruction loss. In our proposal, the reconstruction tasks are conducted in both the time and time-frequency domains, similar to \citep{defossez2022high}. Thus, the reconstruction loss is given by 
\begin{equation}
\label{eq:recons_loss_stage1}
\begin{split}
\mathcal{L}_{\mathrm{recons}} & = \| x_{\mathrm{LF}} - \hat{x}_{\mathrm{LF}} \|_2^2 + \| x_{\mathrm{HF}} - \hat{x}_{\mathrm{HF}} \|_2 \\
                              & + \| u_{\mathrm{LF}} - \hat{u}_{\mathrm{LF}} \|_2^2 + \| u_{\mathrm{HF}} - \hat{u}_{\mathrm{HF}} \|_2^2,
\end{split}
\end{equation}
where $u_{\mathrm{[\cdot]}}$ is equal to $\mathcal{P}_{\mathrm{[\cdot]}}(\mathrm{STFT}(x))$, $\hat{u}$ is the reconstruction of $u$, $x_{\mathrm{[\cdot]}}$ and $\hat{x}_{\mathrm{[\cdot]}}$ are obtained by applying ISTFT to $u_{\mathrm{[\cdot]}}$ and $\hat{u}_{\mathrm{[\cdot]}}$ respectively.

The total training objective for stage~1 becomes:
\begin{equation}
\label{eq:total_loss_stage1}
\mathcal{L}_{\mathrm{VQ}} = \mathcal{L}_{\mathrm{codebook}} + \mathcal{L}_{\mathrm{recons}}.
\end{equation}


\subsection{Stage 2: Prior Learning}

In stage 2, the encoder, decoder, and codebook are frozen and a model is trained on the pre-trained discrete tokens to learn the prior. Inspired by MaskGIT, a bidirectional transformer is used as the prior model. However, the original form of MaskGIT cannot be applied, as there are two different modalities for the tokens, \textit{i.e.,} LF and HF. Our proposal suggests an approach to overcome this problem.

\paragraph{Prior Model Training}
An input can now be represented in terms of the codebook-indices of the discrete tokens and is equivalent to a sequence $s \in \{ 0, 1, \cdots, K-1 \}^{h \times w}$ where, recall, $h$ and $w$ denote height and width of $z_q$. More precisely, each element $s_{ij}$ of such sequence is given by
\begin{equation}
\label{eq:element_of_s}
s_{ij} = k, \:\:\: \textrm{whenever} \:\:\: (z_q)_{ij} = z_k.
\end{equation}

\begin{figure}[]
\centering
\includegraphics[width=0.75\linewidth]{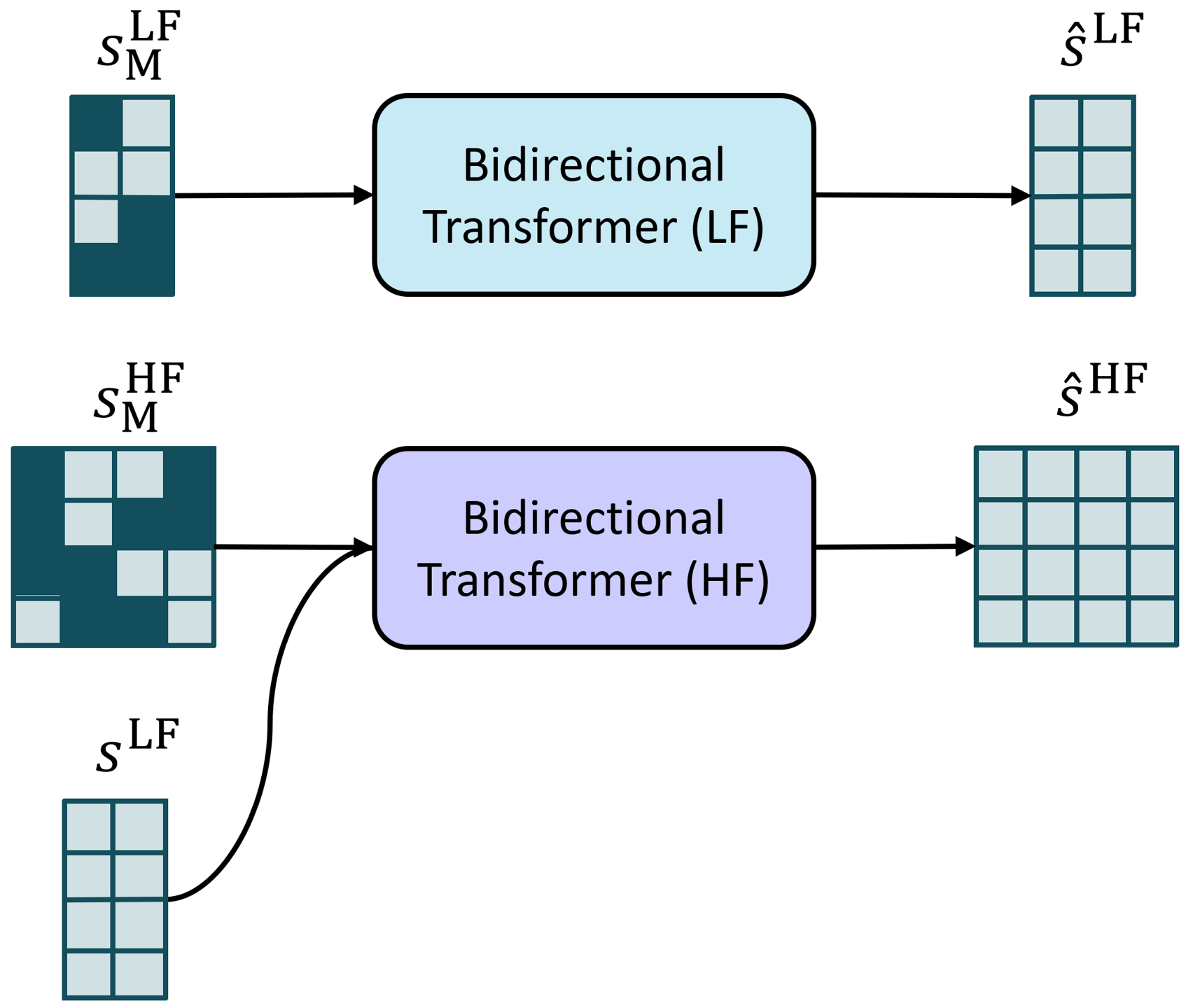}
\caption{Overview of the prior model training (stage 2). The dark green block represents the \texttt{[MASK]} token.}
\label{fig:maksgit_training}
\end{figure}

In the naive MaskGIT, the prior $p(s)$ is modelled by $\sum_{\mathrm{M}}{p(s|s_\mathrm{M})p(s_\mathrm{M})}$, where $s_\mathrm{M}$ is the masked sequence. During training, a random subset of $s$ is replaced by a special \texttt{[MASK]} token to produce $s_\mathrm{M}$. 
In our proposal, we model the prior\footnote{Full derivation is available in Appendix \ref{appendix:full derivation of the joint probability}.} by $p({s}^{\mathrm{LF}}, {s}^{\mathrm{HF}}) \approx \sum_{\mathrm{M}} p({s}^{\mathrm{HF}} | {s}^{\mathrm{LF}}, s_{\mathrm{M}}^{\mathrm{HF}}) p({s}^{\mathrm{LF}} | s_{\mathrm{M}}^{\mathrm{LF}}) p(s_\mathrm{M}^{\mathrm{LF}}) p(s_\mathrm{M}^{\mathrm{HF}})$.
The overview of the prior model training is presented in Fig.~\ref{fig:maksgit_training}, where $\hat{s}$ denotes the predicted $s$. Two bidirectional transformers are used: one for $s^{\mathrm{LF}}$ and the other for $s^{\mathrm{HF}}$; $p(\hat{s}^{\mathrm{LF}} | s_{\mathrm{M}}^{\mathrm{LF}})$ is modelled by the LF bidirectional transformer and $p(\hat{s}^{\mathrm{HF}} | {s}^{\mathrm{LF}}, s_{\mathrm{M}}^{\mathrm{HF}})$ is modelled by the HF bidirectional transformer. 

The training objective is to minimize the negative log-likelihood of the masked tokens, that is, 
\begin{equation}
\begin{split}
\label{eq:prior_model_training_loss}
\mathcal{L}_{\mathrm{mask}} = - \mathbb{E}_{s} [ & \log p_{\theta}({s}^{\mathrm{LF}} | s_{\mathrm{M}}^{\mathrm{LF}}) + \\ & \log p_{\phi}({s}^{\mathrm{HF}} | {s}^{\mathrm{LF}}, s_{\mathrm{M}}^{\mathrm{HF}}) ],
\end{split}
\end{equation}
where $\theta$ and $\phi$ represent the parameters of the LF and HF bidirectional transformers, respectively.


\paragraph{Iterative Decoding}

\begin{figure}[]
\centering
\includegraphics[width=0.99\linewidth]{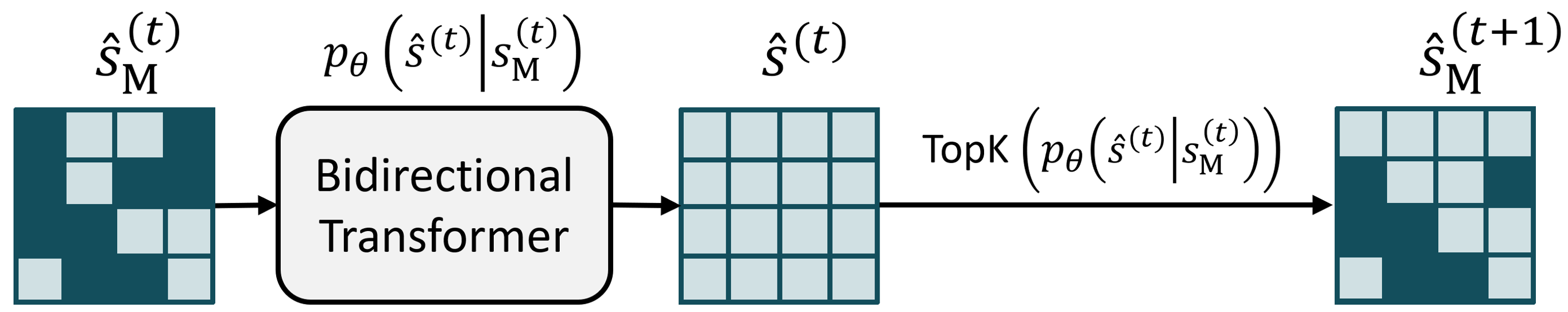}
\caption{Overview of MaskGIT's iterative decoding. The \texttt{TopK} operation is equivalent to \texttt{torch.topk} from PyTorch \citep{NEURIPS2019_9015}. \texttt{TopK} returns \textit{indices} of the $\ell$ largest elements of a given input.}
\label{fig:iterative_decoding}
\end{figure}

In theory, the prior model can predict all the \texttt{[MASK]} tokens and generate a synthetic sample in a single step. However, challenges are encountered if one proceeds this way. Thus, \cite{chang2022maskgit} proposed to predict the \texttt{[MASK]} tokens in iterative multiple steps, \textit{i.e.,} iterative decoding. The overview of MaskGIT's iterative decoding is presented in Fig.~\ref{fig:iterative_decoding}. The decoding process goes from $t=0$ to $T$.
To generate a synthetic sample, we start from a sequence that entirely consists of the \texttt{[MASK]} token indices, which we denote by $s_\mathrm{M}^{(0)}$. We then make predictions for all the \texttt{[MASK]} tokens $p_{\theta}\left( \hat{s}_{ij}^{(t)} | {s}^{(t)}_{\mathrm{M}} \right)$, for $t>0$ and any $i,j$. A number of $\hat{s}_{ij}$ with the largest probabilities are selected to form $\hat{s}_{\mathrm{M}}^{(t+1)}$. This number is determined by a mask scheduling function. Additionally, \cite{chang2022maskgit} used temperature annealing to encourage sample diversity. We use the cosine mask scheduling function and the temperature annealing following \citep{chang2022maskgit}.

\begin{figure*}[t!]
\centering
\includegraphics[width=0.95\linewidth]{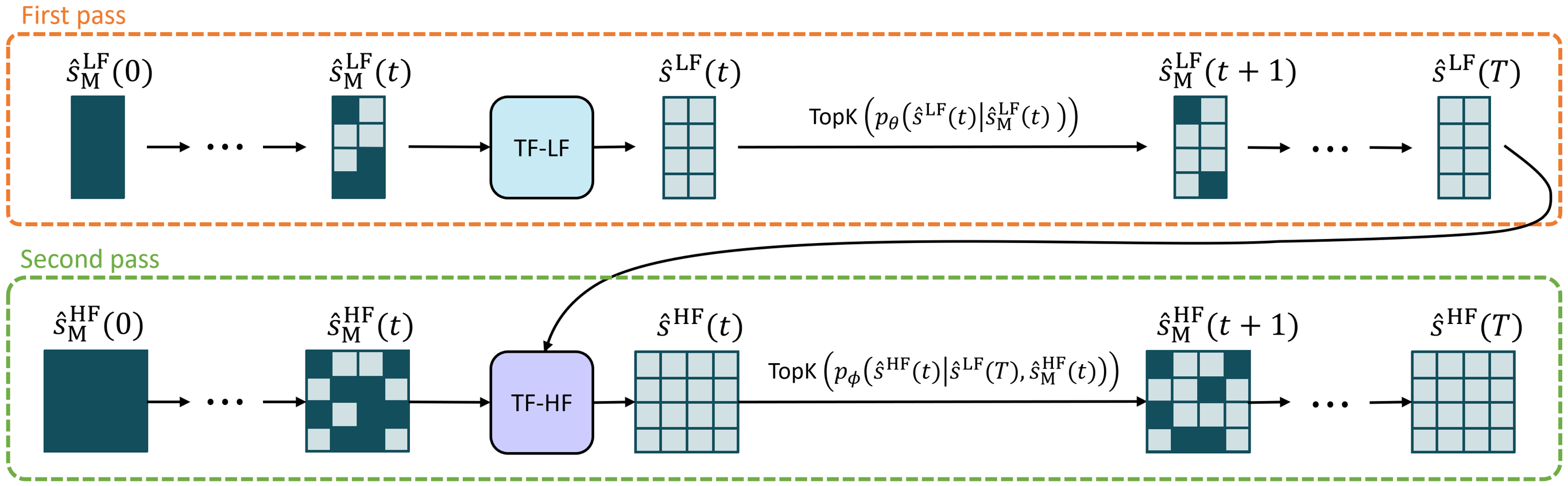}
\caption{Overview of the proposed iterative decoding. It is designed to maximize both $p(\hat{s}^{\mathrm{LF}} | \cdot)$ and $p(\hat{s}^{\mathrm{HF}} | \cdot)$. TF-LF and TF-HF denote the LF and HF bidirectional transformers, respectively.}
\label{fig:two_staged_maskgit}
\end{figure*}

However, because we have two types of $s$, that is, $s^{\mathrm{LF}}$ and $s^{\mathrm{HF}}$, the decoding iteration requires two passes. The overview of the proposed double-pass iterative decoding is presented in Fig.~\ref{fig:two_staged_maskgit}, motivated by Jukebox's ancestral sampling \citep{dhariwal2020jukebox}. In the first pass, $\hat{s}^{\mathrm{LF}}(T)$ is decoded starting from ${s}^{\mathrm{LF}}_{\mathrm{M}}(0)$. In the second pass, 
$\hat{s}^{\mathrm{HF}}(T)$ is decoded starting from ${s}^{\mathrm{HF}}_{\mathrm{M}}(0)$, using the conditional probability dependent on $\hat{s}^{\mathrm{LF}}(T)$ computed in the first pass.
The second pass mimics the training objective $p_{\phi}(s^{\mathrm{HF}} | s^{\mathrm{LF}}, s^{\mathrm{HF}}_{\mathrm{M}} )$ by $p_{\phi}(\hat{s}^{\mathrm{HF}}(t) | \hat{s}^{\mathrm{LF}}(T), \hat{s}^{\mathrm{HF}}_{\mathrm{M}}(t) )$ by assuming $s^{\mathrm{LF}} \approx \hat{s}^{\mathrm{LF}}(T)$ and $s_\mathrm{M}^{\mathrm{HF}} \approx \hat{s}_\mathrm{M}^{\mathrm{HF}}(t)$. To better ensure the assumption, the stochastic sampling \citep{lee2022autoregressive} is used to stochastically sample $s^{\mathrm{LF}}$ and $s^{\mathrm{HF}}$ during training. It can reduce the discrepancy between predictions in training and inference. With the stochastic sampling, $z_q$ is sampled as $z_q \sim \sigma(- \| z_{ij} - z_k \|)$ where $\sigma$ denotes the softmax distribution.

\begin{figure}[]
\centering
\includegraphics[width=1.\linewidth]{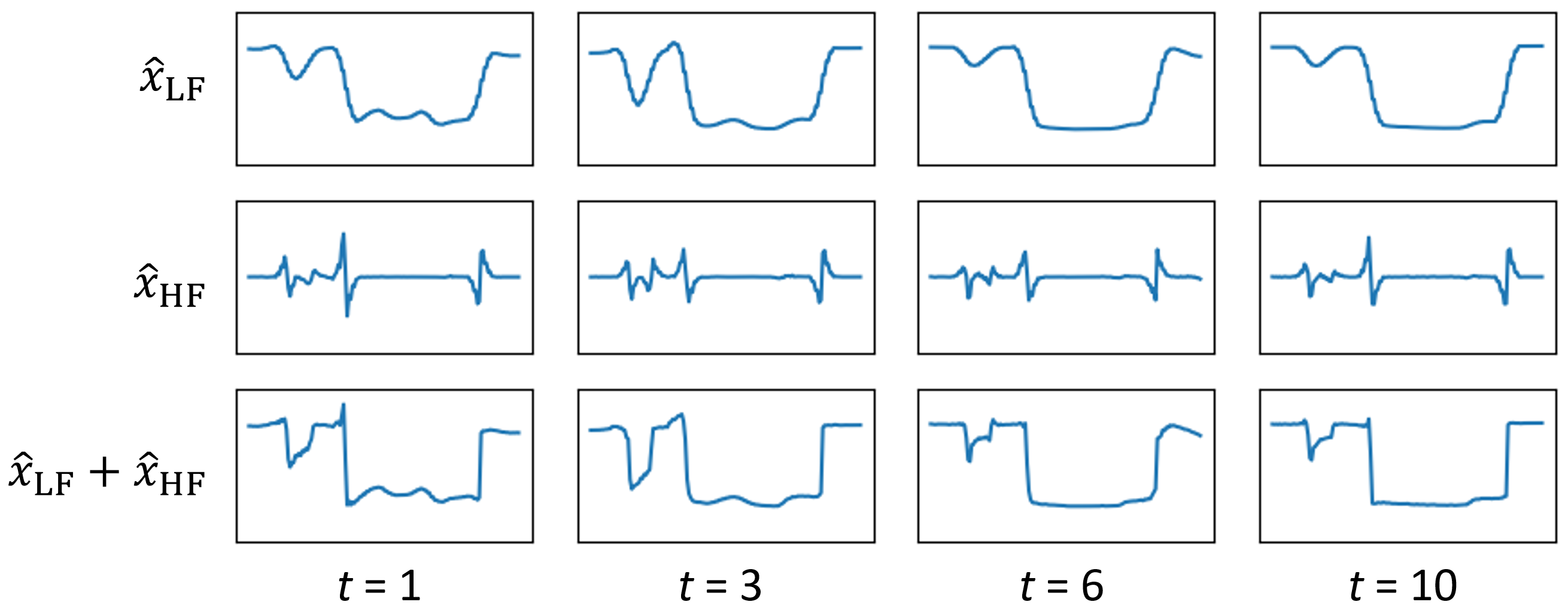}
\caption{Example of the proposed iterative decoding process. Each column shows decoded time series of $\hat{s}^\mathrm{LF}(t)$ and $\hat{s}^\mathrm{HF}(t)$. The dataset of interest is Wafer.}
\label{fig:example_proposed_iterative_decoding}
\end{figure}

To generate synthetic time series, $\hat{s}^{\mathrm{LF}}(T)$ and $\hat{s}^{\mathrm{HF}}(T)$ are mapped back to their corresponding discrete tokens $z_q^{\mathrm{LF}}$ and $z_q^{\mathrm{HF}}$.  Afterwards, the discrete tokens are decoded to $\hat{x}_{\mathrm{LF}}$ and $\hat{x}_{\mathrm{HF}}$, respectively. The complete synthetic time series is obtained by $\hat{x} = \hat{x}_{\mathrm{LF}} + \hat{x}_{\mathrm{HF}}$. An illustrative example of the proposed iterative decoding is presented in Fig.~\ref{fig:example_proposed_iterative_decoding}.

\paragraph{Learning to Sample both Unconditionally and Class-Conditionally}
We train our model such that it can generate synthetic time series both unconditionally and class-conditionally. 
It is achieved by appending a class token index $c$ to $s$ similar to how a class token is appended in the Vision Transformer model \citep{dosovitskiy2020image}. Then $c$ is stochastically sampled by $c \sim \{ $\O$, c_0, c_1, .. \}$ where $\O$ indicates no-class (for unconditional sampling) and $c_i$ indicates a certain class (for class-conditional sampling). This approach is adopted from the classifier-free guidance technique \citep{ho2022classifier}. The probability of sampling $\O$ is given by $p_{\mathrm{uncond}}$, so the probability of sampling one of the $c_i$ is given by $1 - p_{\mathrm{uncond}}$. Following \citep{ho2022classifier}, $p_{\mathrm{uncond}}$ is set to 0.2. 
\cite{gafni2022make} proposes guided sampling for a VQ-based text-to-image model. We slightly modify it and propose guided class-conditional sampling: $p_{g}(\hat{s} | \hat{s}_\mathrm{M}, c_i) = p(\hat{s} | \hat{s}_\mathrm{M}, \O) + \alpha_g \left(p(\hat{s} | \hat{s}_\mathrm{M}, c_i) - p(\hat{s} | \hat{s}_\mathrm{M}, \O)\right)$
where $\alpha_g$ is the guidance scale and the subscript $g$ denotes the guided class-conditional sampling. When $\alpha_g$ is larger than 1, class-conditional sampling complies with class information better.

\section{EXPERIMENTS\footnote{Full results are available in Appendix \ref{appendix:full results}.}}

\subsection{Evaluation Metrics}
There are mainly three metrics used in the experiments: IS, FID score, and CAS. IS measures the quality of synthetic samples by using the entropy of the distribution of labels predicted by a pretrained model -- \textit{i.e.,}, Inception v3 \citep{szegedy2015going} for IMG and FCN for TSG in this work -- and evenness of the predictions across all labels. IS ranges from 1 to the number of classes. Better quality corresponds to higher IS. Unlike IS, FID score measures the quality by comparing distributions of generated samples and real samples. Its lowest score is 0 and it has no upper boundary. Better quality corresponds to lower FID. CAS is used to assess the quality of class-conditionally generated samples. It involves first training a classifier on synthetic samples -- \textit{i.e.,} ResNet-50 \citep{he2016deep} for IMG and FCN for TSG in this work --  and then testing on real data, measuring the resulting accuracy.

\subsection{Experimental Setup}

All datasets from the UCR archive are used in the experiments. Each dataset is normalized such that it has zero mean and unit variance, following \citep{franceschi2019unsupervised}. 
For our encoder and decoder, those from VQ-VAE are adopted.
Our implementation for VQ uses the library from \citep{vq_github}.
The implementation of the prior models, \textit{i.e.,} bidirectional transformers, is taken from \citep{x_transformer_github}.
Further details on the parameter choices and implementation are available in Appendix \ref{appendix:implementation details}.

\subsection{Unconditional Time Series Generation}
\begin{figure}[h]
\includegraphics[width=\linewidth]{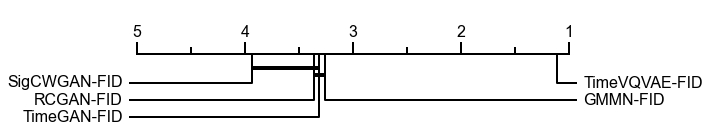}
\includegraphics[width=\linewidth]{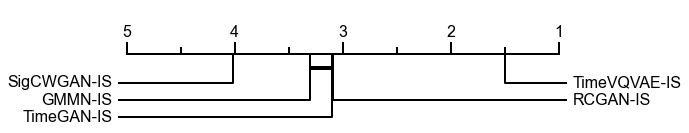}
\caption{Critical difference diagram of the unconditional sampling performances in terms of FID and IS.
The FID scores are multiplied by a factor of -1 in order to rank them in ascending order. Therefore, a higher rank indicates a lower FID score overall. For IS, a higher rank indicates a higher IS overall.}
\label{fig:main_comparison}
\end{figure}

\begin{figure*}[t!]
\centering
\includegraphics[width=0.85\linewidth]{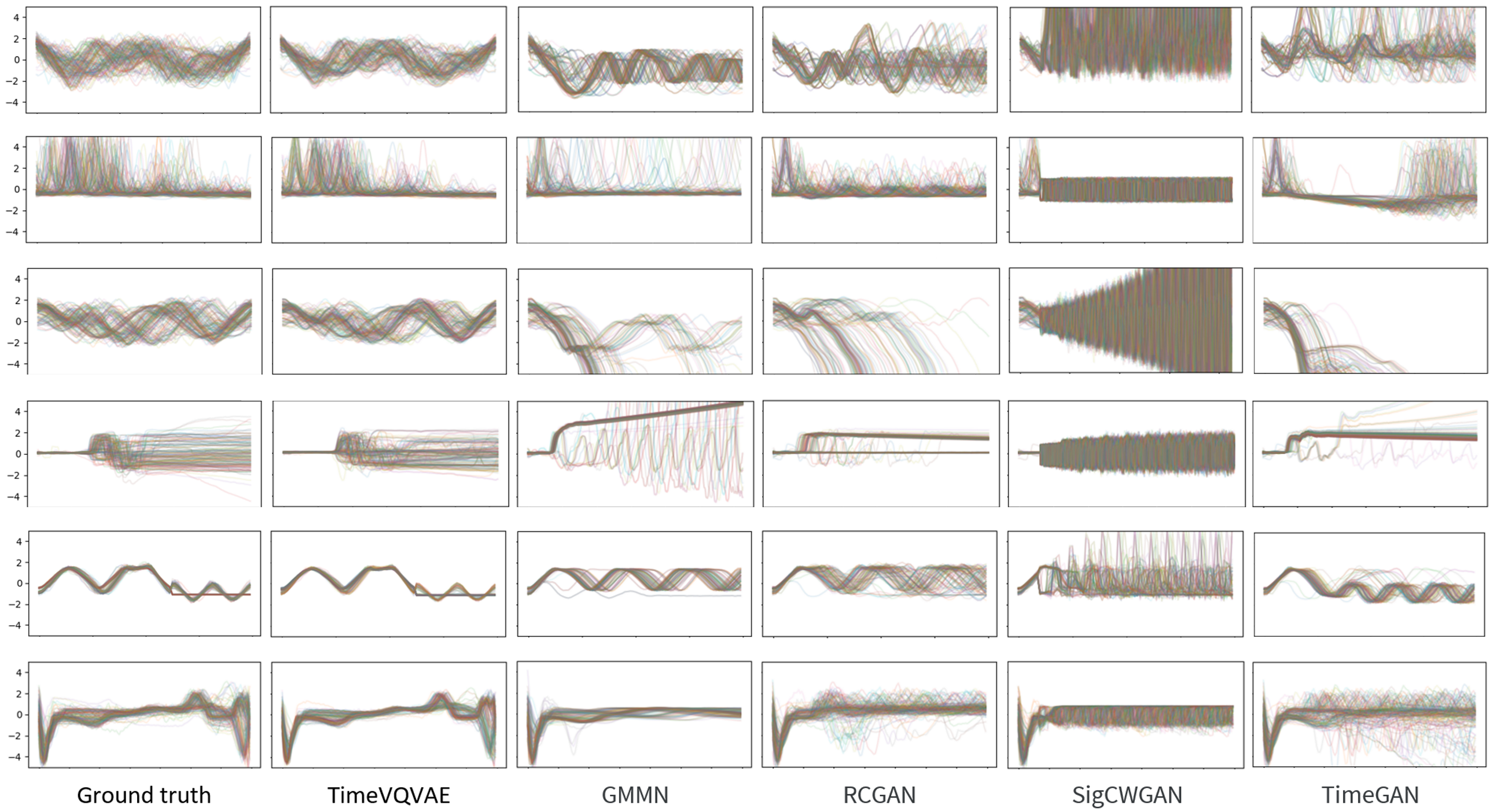}
\caption{Visual inspection of generated samples by the different methods. 
Each row presents each of the following UCR datasets: MixedShapesSmallTrain, InsectWingbeatSound, Yoga, EOGHorizontalSignal, and ECG5000.
}
\label{fig:visual_inspection_generated_samples}
\end{figure*}

GMMN, RCGAN, TimeGAN, SigCWGAN, and TimeVQVAE are compared for the unconditional sampling experiment in terms of FID and IS in Fig.~\ref{fig:main_comparison}. No code is available for TTS-CGAN at the time of writing, thus, it has not been included in the comparative study.
The implementations of the competing models are taken from \citep{sigcwgan_github}.
Among them, RCGAN and TimeGAN are the two most compared methods in the TSG literature. The model sizes are adjusted to be comparable to each other. Although TSGAN reports the FID scores on the UCR datasets, the reported FID scores are far off the normal range of the FID scores, that is, the scores are too high. Thus, due to the concern of a potential miscalculation, TSGAN's reported FID scores are not used here. Fig.~\ref{fig:main_comparison} shows that TimeVQVAE considerably outperforms its competing methods. Fig.~\ref{fig:visual_inspection_generated_samples} presents visualization of generated samples by the different methods.
It is noticeable that the competing methods suffer from the limitation of RNN, that is, RNN typically has difficulties with temporal consistency between two distant timesteps. TimeVQVAE does not have the problem because all time steps are globally attended by the transformer's self-attention mechanism.

\subsection{Class-conditional Time Series Generation}
\begin{figure}[H]
\includegraphics[width=0.96\linewidth]{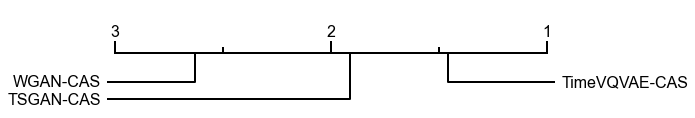}
\caption{Critical difference diagram of the class-conditional performances in terms of CAS. A higher rank indicates higher CAS overall.}
\label{fig:CAS_score_comparison}
\end{figure}

\cite{smith2020conditional} reports CAS of WGAN and TSGAN on 70 subset datasets of the UCR archive. In Fig.~\ref{fig:CAS_score_comparison}, WGAN, TSGAN, and TimeVQVAE are compared for the class-conditional sampling in terms of CAS. $\alpha_g$, in this experiment, is set to 1.
It indicates the synthetic sample distributions generated by TimeVQVAE is closer to the real distribution than WGAN and TSGAN.
There is another advantage of TimeVQVAE. \cite{smith2020conditional} implements the class-conditional sampling by training the same model $N$ times on $N$ different subsets of a dataset according to $N$ different classes. TimeVQVAE, on the other hand, is trained for unconditional and class-conditional sampling at the same time.

\subsection{Ablation Study}
\label{sect:ablation_study}

\begin{figure}[t]
\centering
\includegraphics[width=\linewidth]{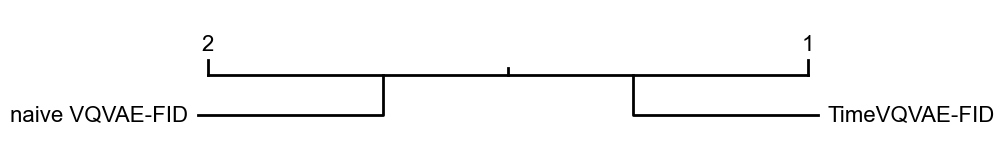}
\includegraphics[width=0.97\linewidth]{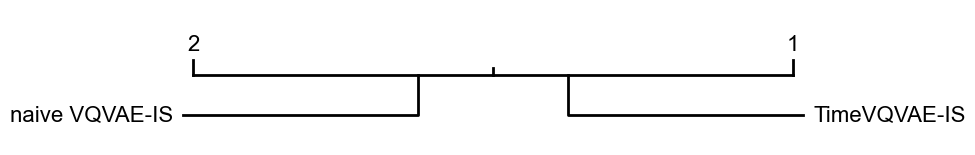}
\caption{Performance comparison between the naive VQ-VAE and TimeVQVAE.}
\label{fig:ablation_naiveVQVAE_TimeVQVAE}
\end{figure}

\paragraph{Naive VQ-VAE vs TimeVQVAE} 
Our proposed VQ modeling adds several novelties on top of VQ-VAE. 
Two cases are experimented: 1) naive VQ-VAE for stage~1 with the MaskGIT's approach for stage~2, 2) TimeVQVAE. The naive VQ-VAE simply takes time series instead of images. Fig. \ref{fig:ablation_naiveVQVAE_TimeVQVAE} shows the considerable performance advantage of TimeVQVAE over its naive form. 

\begin{figure}[t]
\centering
\includegraphics[width=\linewidth]{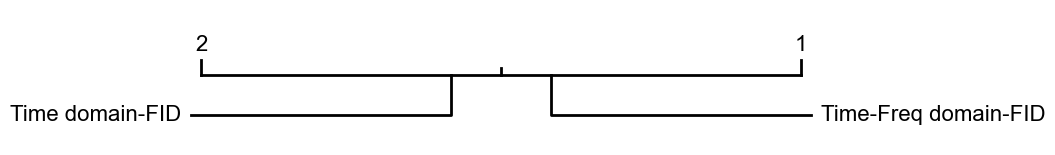}
\includegraphics[width=0.98\linewidth]{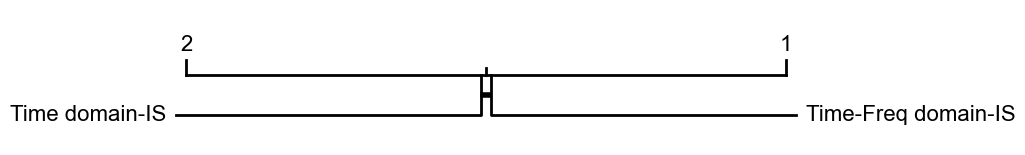}
\caption{Performance comparison between VQ modeling in the time domain and the time-frequency domain.}
\label{fig:ablation_time_vs_timefreq}
\end{figure}

\paragraph{VQ in Time and Time-frequency Domains} 
The VQ modeling can be done in the time domain -- the naive VQ-VAE for stage~1 -- or in the time-frequency domain as proposed. Fig. \ref{fig:ablation_time_vs_timefreq} shows that the VQ modeling in the time-frequency domain is more beneficial. To sorely compare the domain difference, the LF-HF separation is not used for the latter case and both cases have the same downsampling rate.

\begin{figure}[t]
\centering
\includegraphics[width=0.9\linewidth]{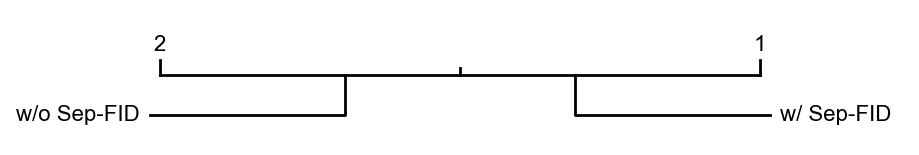}
\includegraphics[width=0.9\linewidth]{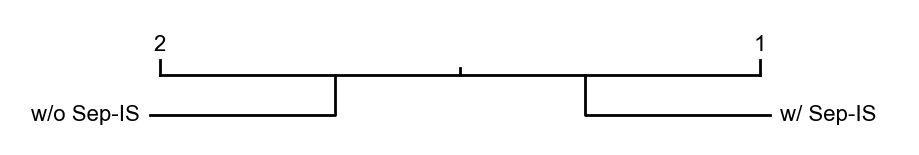}
\caption{Ablation study results with respect to the LF-HF separation. w/ Sep and w/o Sep denote with and without the LF-HF separation, respectively.}
\label{fig:ablation_LF_HF_disentanglement}
\end{figure}

\paragraph{Separation of LF and HF Latent Spaces}
The LF and HF latent spaces are separately learned in TimeVQVAE. Fig.~\ref{fig:ablation_LF_HF_disentanglement} shows the positive effects of the LF-HF separation. 

\begin{figure}[t!]
\centering
\includegraphics[width=1.\linewidth]{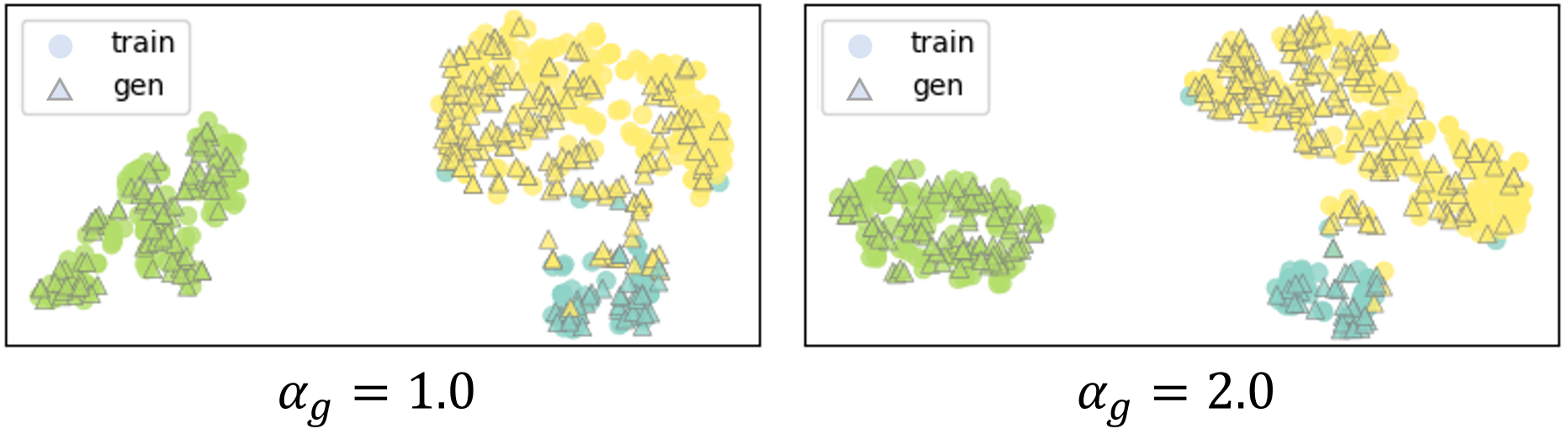}
\caption{2-dimensional t-SNE mapping of the pretrained FCN's representations of real samples (train) and generated samples (gen). The used dataset is StarLightCurves. The different colors denote different classes.}
\label{fig:guided_class_sampling_ablation}
\end{figure}

\paragraph{Guided Class-conditional Sampling}
Fig. \ref{fig:guided_class_sampling_ablation} shows that the higher $\alpha_g$ results in the clearer class-boundaries of the generated samples.

\begin{figure}[t!]
\centering
\includegraphics[width=\linewidth]{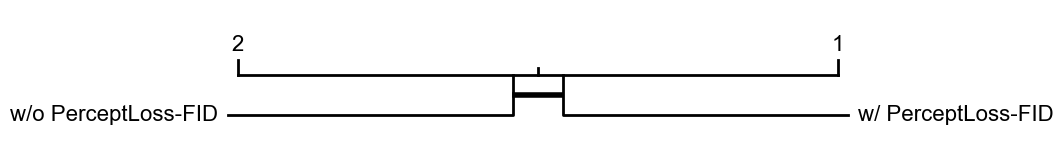}
\includegraphics[width=0.97\linewidth]{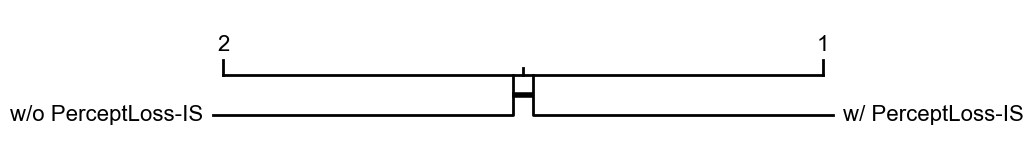}
\caption{Ablation study results with respect to the perceptual loss (PerceptLoss).}
\label{fig:ablation_perceptual_loss}
\end{figure}

\paragraph{Perceptual Loss}
In VQ-GAN, the GAN and perceptual losses are additionally used from VQ-VAE. Those losses help generate perceptually-pleasing images. 
We experiment with the perceptual loss function proposed by \cite{dosovitskiy2016generating}, $\| C(\hat{x}) - C(x) \|_2^2$ where $C$ denotes a pretrained model for feature extraction and $x$ and $\hat{x}$ are real and reconstructed samples, respectively. In our experiments, the pretrained FCN model is used as $C$. The effects of the perceptual loss are presented in Fig.~\ref{fig:ablation_perceptual_loss}. A minor yet positive performance gain is achieved, similarly to the IMG studies \citep{johnson2016perceptual}.

\section{LIMITATIONS}
Because TimeVQVAE incorporates the Fourier Transform, it has difficulties with digital signal-like time series such as the Earthquakes and Computers datasets from the UCR archive. This problem, perhaps, could be overcome by employing the Wavelet Transform. It has the advantage of better locality in the time and frequency domains than STFT. Also, MaskGIT has some drawbacks \citep{lezama2022improved}. Its token selection is made independently for each token and is based on predicted confidence by the generator, which can be prone to a modeling error. Moreover, the selected tokens are not correctable in a later iteration. Resolving the above-mentioned limitations could lead to better TSG.

\section{CONCLUSION}
Motivated by the successes in the IMG literature, we propose TimeVQVAE for TSG by leveraging VQ-VAE.
Our experiments show that TimeVQVAE outperforms its competing methods in both unconditional and class-conditional sampling, plus it can be trained for both tasks at the same time, which greatly increases efficiency. Moreover, positive performance gains are shown by adopting VQ modeling in the time-frequency domain, carrying out LF-HF latent space separation, and including the guided class-conditional sampling and the perceptual loss.

\subsubsection*{Acknowledgements}
We would like to thank the Norwegian Research Council for funding the Machine Learning for Irregular Time Series (ML4ITS) project (312062). This funding directly supported this research. We also would like to thank Prof. Eamonn Keogh and all the people who have contributed to the UCR time series classification archive for their selfless work.





\bibliography{reference}

\appendix
\onecolumn

\section{FULL DERIVATION OF THE JOINT PROBABILITY OF $s^{\mathrm{LF}}$ AND $s^{\mathrm{HF}}$}
\label{appendix:full derivation of the joint probability}

\begin{equation}
\label{eq:p_s}
\begin{split}
& p(s^{\mathrm{LF}}, s^{\mathrm{HF}}) \\
&= \sum_{\mathrm{M}} p(s^{\mathrm{LF}}, s^{\mathrm{HF}}, s^{\mathrm{LF}}_{\mathrm{M}}, s^{\mathrm{HF}}_{\mathrm{M}}) \\
&= \sum_{\mathrm{M}} p(s^{\mathrm{HF}} | s^{\mathrm{LF}}, s^{\mathrm{LF}}_{\mathrm{M}}, s^{\mathrm{HF}}_{\mathrm{M}}) p(s^{\mathrm{LF}}, s^{\mathrm{LF}}_{\mathrm{M}}, s^{\mathrm{HF}}_{\mathrm{M}}) \\
&= \sum_{\mathrm{M}} p(s^{\mathrm{HF}} | s^{\mathrm{LF}}, s^{\mathrm{LF}}_{\mathrm{M}}, s^{\mathrm{HF}}_{\mathrm{M}}) p(s^{\mathrm{LF}} | s^{\mathrm{LF}}_{\mathrm{M}}, s^{\mathrm{HF}}_{\mathrm{M}}) p(s^{\mathrm{LF}}_{\mathrm{M}}, s^{\mathrm{HF}}_{\mathrm{M}}) \\
&\approx \sum_{\mathrm{M}} p(s^{\mathrm{HF}} | s^{\mathrm{LF}}, s^{\mathrm{HF}}_{\mathrm{M}}) p(s^{\mathrm{LF}} | s^{\mathrm{LF}}_{\mathrm{M}}, s^{\mathrm{HF}}_{\mathrm{M}}) p(s^{\mathrm{LF}}_{\mathrm{M}}, s^{\mathrm{HF}}_{\mathrm{M}}) 
\end{split}
\end{equation}

Recall that by HF and LF we mean \textit{high frequency} and \textit{low frequency}, respectively. In the above equation, $p(s^{\mathrm{HF}} | s^{\mathrm{LF}}, s^{\mathrm{LF}}_{\mathrm{M}}, s^{\mathrm{HF}}_{\mathrm{M}})$ is replaced by $p(s^{\mathrm{HF}} | s^{\mathrm{LF}}, s^{\mathrm{HF}}_{\mathrm{M}})$ because $s^{\mathrm{LF}}_{\mathrm{M}}$ is a subset of $s^{\mathrm{LF}}$ and information of the masked tokens is provided by $s^{\mathrm{HF}}_{\mathrm{M}}$. 
The equation can be further simplified with the following independence assumptions -- $p(s^{\mathrm{LF}} | s_\mathrm{M}^{\mathrm{HF}}) = p(s^{\mathrm{LF}})$ and $p(s_\mathrm{M}^\mathrm{LF}, s_\mathrm{M}^\mathrm{HF}) = p(s_\mathrm{M}^\mathrm{LF}) p(s_\mathrm{M}^\mathrm{HF})$. 
The second and third terms of the last equation in Eq. \eqref{eq:p_s} are simplified with the assumptions. We show the simplification of the second term and third term in order. Then, we finally show the simplified equation of Eq. \eqref{eq:p_s}.

The second term is simplified as:
\begin{equation}
\begin{split}
& p(s^{\mathrm{LF}} | s^{\mathrm{LF}}_{\mathrm{M}}, s^{\mathrm{HF}}_{\mathrm{M}}) \\
&= \frac{p(s_\mathrm{M}^\mathrm{LF}, s_\mathrm{M}^\mathrm{HF} | s^{\mathrm{LF}}) p(s^{\mathrm{LF}})}{p(s_\mathrm{M}^\mathrm{LF}, s_\mathrm{M}^\mathrm{HF})} \\
&= \frac{p(s_\mathrm{M}^\mathrm{LF} | s^{\mathrm{LF}}) p(s_\mathrm{M}^\mathrm{HF} | s^{\mathrm{LF}}) p(s^{\mathrm{LF}})}{p(s_\mathrm{M}^\mathrm{LF}) p(s_\mathrm{M}^\mathrm{HF})} \\
&= \frac{p(s_\mathrm{M}^\mathrm{LF} | s^{\mathrm{LF}}) \frac{p(s^{\mathrm{LF}} | s_\mathrm{M}^\mathrm{HF}) p(s_\mathrm{M}^\mathrm{HF})}{p(s^\mathrm{LF})} p(s^{\mathrm{LF}})}{p(s_\mathrm{M}^\mathrm{LF}) p(s_\mathrm{M}^\mathrm{HF})} \\
&= \frac{p(s_\mathrm{M}^\mathrm{LF} | s^{\mathrm{LF}}) \frac{p(s^{\mathrm{LF}}) p(s_\mathrm{M}^\mathrm{HF})}{p(s^\mathrm{LF})} p(s^{\mathrm{LF}})}{p(s_\mathrm{M}^\mathrm{LF}) p(s_\mathrm{M}^\mathrm{HF})} \\
&= \frac{p(s_\mathrm{M}^\mathrm{LF} | s^{\mathrm{LF}}) p(s_\mathrm{M}^\mathrm{HF}) p(s^{\mathrm{LF}})}{p(s_\mathrm{M}^\mathrm{LF}) p(s_\mathrm{M}^\mathrm{HF})} \\
&= \frac{p(s_\mathrm{M}^\mathrm{LF} | s^{\mathrm{LF}}) p(s^{\mathrm{LF}})}{p(s_\mathrm{M}^\mathrm{LF})} \\
&= p(s^{\mathrm{LF}} | s_\mathrm{M}^\mathrm{LF})
\end{split}
\end{equation}

The third term is simplified as:
\begin{equation}
p(s^{\mathrm{LF}}_{\mathrm{M}}, s^{\mathrm{HF}}_{\mathrm{M}}) = p(s^{\mathrm{LF}}_{\mathrm{M}}) p(s^{\mathrm{HF}}_{\mathrm{M}})
\end{equation}

Finally, Eq. \eqref{eq:p_s} is simplified to:
\begin{equation}
\label{eq:p_s_simplified}
\begin{split}
p(s^{\mathrm{LF}}, s^{\mathrm{HF}}) = \sum_{\mathrm{M}} p(s^{\mathrm{HF}} | s^{\mathrm{LF}}, s^{\mathrm{HF}}_{\mathrm{M}}) p(s^{\mathrm{LF}} | s^{\mathrm{LF}}_{\mathrm{M}}) p(s^{\mathrm{LF}}_{\mathrm{M}}) p(s^{\mathrm{HF}}_{\mathrm{M}})
\end{split}
\end{equation}

\section{PRETRAINING SETUP FOR THE SUPERVISED FCN MODEL}
\label{appendix:pretraining setup for the supervised fcn model}
The FCN model consists of a few convolutional blocks with the global average pooling (GAP), linear, and softmax layers followed.  
The source code for the FCN model is from \citep{fcn_reference_github}. Its representation (feature) vector for computing FID is extracted right after the GAP layer. 
AdamW \citep{loshchilov2017decoupled} is used for an optimizer with \{batch size: 256, initial learning rate (LR): 1e-3, LR scheduler: cosine scheduler, weight decay: 1e-5, max epoch: 1,000\}. An open-source code for the pretrained FCN model is available on \url{https://github.com/danelee2601/supervised-FCN}.

\section{IMPLEMENTATION DETAILS}
\label{appendix:implementation details}

\subsection{Datasets}
For the unconditional and class-conditional sampling experiments, 128 datasets, \textit{i.e.,} all the datasets, from the UCR archive are used, on which IS, FID, and CAS are reported.
CAS reported by \cite{smith2020one} is available only for 70 datasets out of 128. The missing scores are left blank in our result table below.

\subsection{Evaluation Metrics: IS, FID, and CAS}
To compute IS and FID, the same number of synthetic samples is unconditionally generated as that of a test set unless the test set has less than 256 samples. 
For datasets with less than 256 test samples, 256 synthetic samples are generated to better ensure the distribution of $\hat{x}~\sim~p_{\theta,\phi}(s^{\mathrm{LF}}, s^{\mathrm{HF}})$, which results in more consistent IS and FID score.

To compute CAS, the same number of synthetic samples per class is generated as that of a training set. If the training set has less than 1,000 samples, 1,000 samples are generated according to the class distribution of the training set to ensure the distribution of $\hat{x}~\sim~p_{\theta,\phi}(s^{\mathrm{LF}}, s^{\mathrm{HF}} | c)$ and to prevent the overfitting problem in the FCN model's training. $c$ denotes the class token index. 
For instance, if a training set has 50 and 150 samples for class~1 and class~2, then 5 times the number of samples of each class are generated -- that is, 250 for class~1 and 750 for class~2 -- so that a total number of generated samples becomes 1,000. An average of 5 CAS from 5 runs is reported.

\subsection{TimeVQVAE}

\paragraph{STFT, ISTFT, and Convolutional Kernel and Stride Sizes} STFT and ISTFT are implemented with \texttt{torch.stft} and \texttt{torch.istft}, respectively. Their main parameter -- \texttt{n\_fft}, size of Fourier transform -- can be any arbitrary integer, but we set it to 8 and we use default parameters for the rest.
The frequency is separated into LF and HF by assigning the bottom rows (with the lowest $\omega$) to LF and the rest to HF.
\texttt{n\_fft} of 8 indicates that the frequency axis $\omega$ has a range of [1, 2, 3, 4, 5] and the length of the temporal dimension becomes half as long as a given time series because \texttt{hop\_length} is set as \texttt{n\_fft/4} by default. 
This setting is chosen due to different temporal lengths of datasets, causing difficulties with selecting the convolutional kernel and stride sizes for downsampling. The encoder has several downsampling layers to compress an input into the latent space. The downsampling rate is determined by the kernel and stride sizes, typically downsampling by 2 with \{kernel\_size:4, stride=2, padding=1\} at each downsampling layer. But a choice of the sizes is difficult because the temporal and frequency axes have different lengths across the different datasets. Therefore, we let the downsampling layers downsample along the temporal axis only, not the frequency axis. Then, the downsampling layer becomes \texttt{nn.Conv2d} with \{kernel\_size=(3,4), stride=(1,2), padding=(1,1)\}.
We have experimentally found that smaller \texttt{n\_fft} such as 4 or 8 leads to better performance. That is associated with compression amount of input data. The higher \texttt{n\_fft} leads to the wider frequency axis and shorter temporal axis, and the compression amount becomes smaller because the downsampling is only applied along the temporal axis. The global consistency of generated samples is typically poorer with the smaller compression amount (\textit{i.e.,} a lower downsampling rate).

\paragraph{Encoder and Decoder} 
The same encoder and decoder architectures from the VQ-VAE paper are used and their implementations are from \citep{encoder_decoder_github}. 
The encoder consists of $n$ downsampling convolutional blocks (Conv2d -- BatchNorm2d -- LeakyReLU), followed by $m$ residual blocks (LeakyReLU -- Conv2d -- BatchNorm2d -- LeakyReLU -- Conv2d). The downsampling convolutional layers are implemented by \texttt{nn.Conv2d(kernel\_size=(3,4), stride=(1,2), padding=(1,1))} -- Note that the downsampling is to be conducted along the temporal axis only. The residual convolutional layers are implemented by \texttt{nn.Conv2d(kernel\_size=(3,3), stride=(1,1), padding=(1,1))}.
The decoder similarly has $m$ residual blocks, followed by $n$ upsampling convolutional blocks. The upsampling convolutional layers are implemented by \texttt{nn.ConvTranspose2d(kernel\_size=(3,4), stride=(1,2), padding=(1,1))}. 
The downsampling rate is determined as $2^n$. We set $n$ such that $z$ has width of around 8 and 32 for the LF and HF encoders and decoders, respectively. We refer to the width as the \textit{downsampled width}. Note that the downsampled width cannot exactly be the specified value unless input length can be expressed as $2^n$. We set the downsampling rate such that the width of $z$ can be as close as to the specified downsampled width.
If a dataset has temporal length shorter than the downsampled width, the downsampling rate is set to 1 -- \textit{i.e.,} no downsampling.
Different sizes of the encoder and decoder are specified in Table \ref{tab:size_enc_dec}. Large size is not considered due to our computational limitation.
The \textit{Base}-sized encoder and decoder are used for the unconditional and class-conditional sampling experiments.

\begin{table}[h]
\caption{Different sizes of the encoder and decoder.}
\label{tab:size_enc_dec}
\centering
\begin{tabular}{l ccc}
\hline
Sizes of the encoder \& decoder   & Small & Base \\
\hline
Hidden dimension size          & 32    & 64    \\
Number of residual blocks & 2     & 4         \\
\hline
\end{tabular}
\end{table}


\paragraph{Trick for Length Match Between $x$ and $\hat{x}$}
When length of $x$ does not follow $2^n$ or is an odd number such as 89, length mismatch between $x$ and $\hat{x}$ occurs. The problem can be easily resolved with the following trick: Adding one additional upsampling layer, followed by linear interpolation in the decoder. \texttt{torch.nn.functional.interpolate(.., size=length(x), mode='linear')} is used for the interpolation in our implementation. If the interpolation is only used without the additional upsampling layer, it leads to loss of high-frequency information in $\hat{x}$.

\paragraph{VQ}
Our implementation for VQ-VAE is from \citep{vq_github}. 
The codebook size $K$ is set to 32 and the code dimension size is the same as the hidden dimension size of the encoder and decoder for both LF and HF. Because the datasets from the UCR archive are less complex and much smaller than an image benchmark dataset such as ImagNet \citep{deng2009imagenet}, the small-sized codebook was found to be sufficient. 
For the codebook-learning loss, we use both the commitment loss and the exponential moving average alternative, following \citep{vq_github}. We use $\beta$ of 1. and the exponential moving average decay rate of 0.8.
The codebook embeddings learned in stage~1 are used to initialize the codebook embeddings in stage~2 to benefit the prior learning. This can be especially helpful when a training set is very small such as Fungi. The dataset, Fungi, has only 18 training samples.

\paragraph{Prior Learning}
The number of iterations, $T$, is set to 10, following \citep{chang2022maskgit}. 
Our implementation for MaskGIT is adopted from \citep{maskgit_github} and implementation for the prior models, \textit{i.e.,} bidirectional transformers, is from \citep{x_transformer_github}. 
Different sizes of the prior model are specified in Table \ref{tab:size_prior_model}. 
The \textit{Base} size is used for the unconditional and class-conditional sampling experiments.
To better stabilize the transformer model, we use Root Mean Square Layer Normalization proposed by \cite{zhang2019root}.

\begin{table}[h]
\caption{Different sizes of the prior model -- bidirectional transformer. Its sizes are referred from \citep{hassani2021escaping}. The feed-forward (FF) ratio denotes a ratio of hidden dimension size to input dimension size. Note that the FF layer augments its input into a larger dimension and back into its original dimension.}
\label{tab:size_prior_model}
\centering
\begin{tabular}{l ccc}
\hline
Size of the transformer  & Small & Base \\
\hline
Hidden dimension size          & 64    &  256       \\
Number of layers   &  2     &  4          \\
Number of heads   &  2     &  2         \\
Feed-forward ratio   &  1     &  1          \\
\hline
\end{tabular}
\end{table}

\paragraph{Optimizer}
\label{sect:optimizer}
The AdamW optimizer is used with \{batch size for stage~1: 128, batch size for stage~2: 256, initial LR: 1e-3, LR scheduler: cosine scheduler, weight decay: 1e-5\}. 
Maximum epochs are \{stage~1: 2,000, stage~2: 10,000\} for the TimeVQVAE model in the unconditional and class-conditional sampling experiments.

\subsection{GMMN, RCGAN, TimeGAN, and SigCWGAN}
        
The implementations of GMMN, RCGAN, TimeGAN, and SigCWGAN from \citep{sigcwgan_github} are used for the unconditional sampling experiments. 
For the above-mentioned methods, AR-FNN (Auto-Regressive Feed-forward Neural Network) proposed by \cite{ni2020conditional} is used for a generator and discriminator. 
To be comparable with TimeVQVAE in terms of a number of trainable parameters, the AR-FNN model is set to have 18 hidden layers and hidden dimension size of 32. The exact number of trainable parameters varies depending on datasets due to different lengths.
The AR-FNN generator requires values at the previous $p$ timesteps to predict the next $q$ timesteps. $p$ is determined as 10\% of input length. The longer $p$ compromises the integrity of data generation and the short $p$ makes the autoregressive model difficult to generate reasonable samples. $q$ is set to 1 for GMMN, RCGAN, and TimeGAN and 3 for SigCWGAN. $q$ does not have to be any larger because data is autoregressively generated at each timestep.
The Adam optimizer \citep{kingma2014adam} is used for both generator and discriminator with \{batch size: 256, initial LR: 2e-4\, max\_epochs: 10,000\}.

\subsection{Training the FCN Model for CAS}
To compute CAS, the FCN model is trained on a set of synthetic samples. To train the FCN model, the AdamW optimizer is used with \{batch~size: 256, initial LR: 1e-3, LR scheduler: cosine scheduler, weight decay: 1e-5, max epochs: 1,000\}. Basically, the same setting as the supervised FCN model is used.

\section{EXPERIMENT DETAILS FOR THE ABLATION STUDIES}
\label{appendix:experiment details for the ablation studies}
The experiments for the ablation studies are conducted in a smaller scale due to a large number of experimental cases. 
The datasets with training set size equal or smaller than 1,000 are used -- Yet, it still sums to 120 datasets; enough to capture the overall ranking. 
The optimizer settings are the same as in the prior section except that the maximum epochs of \{stage~1: 1,000, stage~2: 5,000\} are used.
For the prior model, the \textit{Small}-sized bidirectional model is used for all the ablation studies.

\subsection{Model Details}
There are 4 ablation studies with quantitative experiments. The model details for each ablation study are stated in the following paragraphs. Unless specified differently, the model parameters are the same as in the prior section.

\paragraph{Naive VQ-VAE vs TimeVQVAE}
The \textit{Small}-sized encoder and decoder are used for TimeVQVAE. TimeVQVAE has two sets of the encoder and decoder, but the naive VQ-VAE has one set of them. To be better comparable, size of the naive VQ-VAE's encoder and decoder is set by \{hidden dimension size: 32, number of residual blocks: 4\} and $K$ is set to 64. The downsampled width of the naive VQ-VAE is to be around 8. For the naive VQ-VAE, 2-dimensional convolutional layers are replaced with the 1-dimensional convolutional layers as it takes 1-dimensional time series as an input instead of 2-dimensional time-frequency data.

\paragraph{VQ in Time and Time-frequency Domains} The encoder and decoder with \{hidden dimension size: 32, number of residual blocks: 4\} are used for both cases. Both, also, use the downsampled width of 8 and $K$ of 64. The case, \textit{VQ in Time domain}, is identical to the naive VQ-VAE.

\paragraph{Separation of LF and HF Latent Spaces}
The case without the LF-HF separation uses the encoder and decoder size of \{hidden dimension size: 32, number of residual blocks: 4\}, $K$ of 64, and the downsampled width of around 8. The case without the separation is identical to the case, \textit{VQ in Time-frequency domain}, above. The case with the separation is identical to TimeVQVAE.

\paragraph{Perceptual Loss} TimeVQVAE with the \textit{Small}-sized encoder and decoder is used.

\section{FULL RESULTS}
\label{appendix:full results}

The number of training parameters of TimeVQVAE in Tables \ref{tab:full_result_fid}-\ref{tab:full_result_cas} varies between 0.8~M and 1.6~M for stage~1 (\textit{i.e.,} encoder, codebook, and decoder) depending on the downsampled width, and the number varies between 2.3~M and 2.4~M for stage~2 (\textit{i.e.,} bidirectional transformer). The results for FID, IS, and CAS in Tables \ref{tab:full_result_fid}-\ref{tab:full_result_cas} are reported with mean and standard deviation of the scores over 3 runs.

\begin{center}
\small
\begin{longtable}{l cccccc}
\caption{Full results of the unconditional sampling experiments with FID. The FID score larger than 1,000 is marked as nan because such a large FID score already indicates that the synthetic sample quality is significantly poor.} \\
\label{tab:full_result_fid} \\ 
\hline
Dataset names & GMMN & RCGAN & TimeGAN & SigCWGAN & TimeVQVAE.mean & TimeVQVAE.std \\ \hline
ACSF1 & 74.3 & 98.3 & 81.3 & nan & 27.1 & 0.8 \\ 
Adiac & 98.2 & 54.3 & 45.1 & nan & 7.8 & 0.9 \\ 
AllGestureWiimoteX & 180.7 & nan & nan & nan & 3.8 & 0.7 \\ 
AllGestureWiimoteY & 23.5 & nan & nan & 660.4 & 3.0 & 0.5 \\ 
AllGestureWiimoteZ & nan & 2.8 & 7.9 & 278.9 & 1.8 & 0.2 \\ 
ArrowHead & 8.1 & 15.5 & 22.2 & nan & 2.0 & 0.3 \\ 
Beef & 215.2 & 36 & 18.3 & nan & 1.8 & 0.2 \\ 
BeetleFly & 246.4 & 348.5 & nan & nan & 2.1 & 1.2 \\ 
BirdChicken & 9.9 & 8.5 & 7.6 & 735.5 & 0.3 & 0.1 \\ 
BME & 77.6 & 179.6 & 140.1 & nan & 23.9 & 5.7 \\ 
Car & 101.5 & 579.5 & 186.1 & nan & 2.6 & 0.4 \\ 
CBF & 147.1 & 483.9 & 5.5 & 24.4 & 2.3 & 0.4 \\ 
Chinatown & 27.3 & 70.4 & 59.9 & 25.8 & 5.5 & 0.9 \\ 
ChlorineConcentration & 41.6 & 7.4 & 7.1 & 132.4 & 1.1 & 0.1 \\ 
CinCECGTorso & nan & 172.3 & 57 & nan & 7.7 & 1.5 \\ 
Coffee & 16.2 & 18.1 & nan & nan & 0.3 & 0.2 \\ 
Computers & 29.1 & nan & nan & 19.8 & 3.5 & 0.4 \\ 
CricketX & 115.8 & 404.8 & 23 & 26.9 & 3.0 & 0.5 \\ 
CricketY & 33 & nan & nan & 17.6 & 3.0 & 0.4 \\ 
CricketZ & 60.1 & 105.3 & 14.4 & 19.8 & 4.0 & 0.3 \\ 
Crop & 80.7 & 14.9 & 18.7 & 35.3 & 3.9 & 0.7 \\ 
DiatomSizeReduction & 138.6 & 232.9 & 88.7 & nan & 12.8 & 1.4 \\ 
DistalPhalanxOutlineAgeGroup & 13.9 & 9 & 110.6 & nan & 6.3 & 1.0 \\ 
DistalPhalanxOutlineCorrect & 9.2 & 12 & 16 & 50.4 & 1.8 & 0.3 \\ 
DistalPhalanxTW & 16.8 & 19.2 & 21.3 & nan & 13.2 & 1.5 \\ 
DodgerLoopDay & 56.8 & 429.9 & 30.1 & 14.4 & 1.9 & 0.2 \\ 
DodgerLoopGame & 44.8 & nan & 21.6 & 63.4 & 5.9 & 0.9 \\ 
DodgerLoopWeekend & 9.6 & 273.2 & 16.1 & 14.7 & 6.2 & 1.1 \\ 
Earthquakes & nan & nan & nan & 5.4 & 1.6 & 0.5 \\ 
ECG200 & 3 & 2.8 & 2.7 & 20.6 & 1.9 & 0.3 \\ 
ECG5000 & 26.6 & 4.5 & 35.2 & 55.9 & 0.7 & 0.0 \\ 
ECGFiveDays & 15 & 22.3 & 7.1 & 523.2 & 0.6 & 0.2 \\ 
ElectricDevices & 37.1 & 151.9 & 79.9 & 105.5 & 6.8 & 1.1 \\ 
EOGHorizontalSignal & 287.8 & 47.3 & 95.5 & nan & 3.5 & 0.6 \\ 
EOGVerticalSignal & 238.1 & 457.7 & 100.6 & nan & 6.8 & 1.2 \\ 
EthanolLevel & 19.4 & 15.7 & 18 & nan & 0.2 & 0.2 \\ 
FaceAll & 42.4 & 10.3 & nan & 33 & 4.4 & 0.0 \\ 
FaceFour & 18.6 & 65 & 56.5 & 26.5 & 3.6 & 1.3 \\ 
FacesUCR & 39.7 & 7.3 & 20.6 & 39.7 & 3.2 & 0.4 \\ 
FiftyWords & nan & 27.7 & 81.7 & 821.9 & 9.7 & 1.5 \\ 
Fish & nan & 36.3 & 47.4 & nan & 11.5 & 0.1 \\ 
FordA & 3.6 & 178 & nan & nan & 3.0 & 0.4 \\ 
FordB & nan & 45.6 & nan & nan & 1.4 & 0.6 \\ 
FreezerRegularTrain & 56 & 41.6 & 27.6 & 176.9 & 7.3 & 1.3 \\ 
FreezerSmallTrain & 49.3 & 32 & 42.7 & 109.6 & 9.5 & 0.7 \\ 
Fungi & 48.4 & 86 & 82.9 & nan & 2.1 & 0.5 \\ 
GestureMidAirD1 & nan & nan & nan & nan & 9.9 & 5.1 \\ 
GestureMidAirD2 & nan & nan & 306.8 & nan & 3.8 & 0.5 \\ 
GestureMidAirD3 & 214.7 & nan & nan & nan & 46.4 & 19.1 \\ 
GesturePebbleZ1 & 163.4 & 29.5 & 10.6 & 26.2 & 3.0 & 2.3 \\ 
GesturePebbleZ2 & 16.7 & 350.8 & 22.3 & 49.9 & 3.7 & 1.1 \\ 
GunPoint & 16 & 8.8 & 3.5 & nan & 0.8 & 0.4 \\ 
GunPointAgeSpan & 314.7 & 19.7 & 267.3 & nan & 1.1 & 0.5 \\ 
GunPointMaleVersusFemale & nan & 44.6 & 237.5 & 215.4 & 0.6 & 0.3 \\ 
GunPointOldVersusYoung & 14.1 & nan & 13.2 & nan & 0.3 & 0.1 \\ 
Ham & nan & 43.5 & 27.4 & 427 & 0.6 & 0.2 \\ 
HandOutlines & 9.4 & 3.7 & 1.3 & nan & 0.1 & 0.0 \\ 
Haptics & nan & nan & nan & nan & 2.5 & 0.2 \\ 
Herring & 35 & 3.9 & 75.8 & nan & 0.5 & 0.1 \\ 
HouseTwenty & 26.6 & nan & 38 & 100.3 & 4.8 & 2.7 \\ 
InlineSkate & 96 & 143.3 & 119.5 & nan & 13.3 & 5.0 \\ 
InsectEPGRegularTrain & 7.3 & 28.5 & 0.9 & 0.3 & 2.5 & 1.0 \\ 
InsectEPGSmallTrain & 121.2 & 113.5 & 125.9 & 80.8 & 3.6 & 3.3 \\ 
InsectWingbeatSound & 14.5 & 6.2 & 132.3 & nan & 9.9 & 0.7 \\ 
ItalyPowerDemand & 5.7 & 57.5 & 5.8 & 4.6 & 1.6 & 0.4 \\ 
LargeKitchenAppliances & 8.3 & 47.3 & 31.8 & 281 & 0.9 & 0.3 \\ 
Lightning2 & 19.5 & 8.6 & 71.7 & 162.3 & 1.2 & 0.4 \\ 
Lightning7 & 79.2 & 78.6 & 27 & 92.1 & 1.1 & 0.2 \\ 
Mallat & nan & 11.4 & nan & nan & 1.1 & 0.2 \\ 
Meat & 134.7 & 18.2 & nan & nan & 7.7 & 3.8 \\ 
MedicalImages & 20.9 & 24 & nan & 55.3 & 3.5 & 0.4 \\ 
MelbournePedestrian & 66.6 & 35.5 & 62.5 & 90.1 & 2.4 & 0.9 \\ 
MiddlePhalanxOutlineAgeGroup & 18.7 & 12.6 & 13.8 & 590.2 & 20.0 & 1.6 \\ 
MiddlePhalanxOutlineCorrect & 7.2 & 21.2 & 162.3 & 535.4 & 1.2 & 0.4 \\ 
MiddlePhalanxTW & 62.2 & 27.4 & 19.8 & nan & 3.3 & 0.5 \\ 
MixedShapesRegularTrain & 379 & 412.2 & nan & nan & 34.1 & 6.5 \\ 
MixedShapesSmallTrain & 20.6 & 13.3 & 262.1 & nan & 7.6 & 1.8 \\ 
MoteStrain & 5.6 & 5.9 & 4 & 13.9 & 1.5 & 0.3 \\ 
NonInvasiveFetalECGThorax1 & 438.5 & nan & nan & nan & 16.0 & 2.8 \\ 
NonInvasiveFetalECGThorax2 & 126.8 & 117.9 & 150 & nan & 30.6 & 13.9 \\ 
OliveOil & 9.8 & 9.5 & 9.9 & nan & 1.7 & 0.1 \\ 
OSULeaf & nan & nan & nan & nan & 13.2 & 0.9 \\ 
PhalangesOutlinesCorrect & 1.9 & 3 & 2.3 & 73.4 & 0.7 & 0.1 \\ 
Phoneme & nan & nan & nan & nan & 12.1 & 0.4 \\ 
PickupGestureWiimoteZ & 275.6 & 380.9 & nan & 116 & 2.1 & 0.2 \\ 
PigAirwayPressure & 111.2 & 188.5 & 642.6 & 274.4 & 38.1 & 11.9 \\ 
PigArtPressure & 291.4 & nan & nan & 80.3 & 72.4 & 13.3 \\ 
PigCVP & 107.6 & nan & nan & 343.3 & 57.7 & 9.7 \\ 
PLAID & 797.4 & nan & 335.9 & nan & 552.8 & 474.5 \\ 
Plane & 39.2 & 38.9 & 36.2 & nan & 6.6 & 0.5 \\ 
PowerCons & 18.3 & 15 & 16.5 & 30 & 1.0 & 0.4 \\ 
ProximalPhalanxOutlineAgeGroup & 73.2 & 93.9 & 22.1 & nan & 0.4 & 0.2 \\ 
ProximalPhalanxOutlineCorrect & 3.7 & 1.7 & 13.8 & nan & 0.4 & 0.2 \\ 
ProximalPhalanxTW & 60.8 & 24.2 & 16.2 & nan & 2.8 & 0.4 \\ 
RefrigerationDevices & nan & nan & nan & 9.2 & 16.2 & 3.6 \\ 
Rock & nan & nan & nan & 23.5 & 6.0 & 3.1 \\ 
ScreenType & 32.7 & nan & 30.9 & 28.4 & 6.3 & 0.8 \\ 
SemgHandGenderCh2 & 14.4 & nan & nan & 69.9 & 4.6 & 0.2 \\ 
SemgHandMovementCh2 & 164.6 & 71.8 & 199.1 & 59.4 & 14.1 & 0.3 \\ 
SemgHandSubjectCh2 & 37 & nan & nan & 108.1 & 29.6 & 2.3 \\ 
ShakeGestureWiimoteZ & nan & 42.5 & 22.4 & 116.7 & 2.4 & 0.7 \\ 
ShapeletSim & 8.4 & 2.6 & nan & 0.8 & 9.0 & 2.5 \\ 
ShapesAll & nan & nan & nan & nan & 14.4 & 2.2 \\ 
SmallKitchenAppliances & 19 & nan & 20.8 & 233.1 & 4.7 & 0.4 \\ 
SmoothSubspace & nan & 7.7 & 14.1 & 8.2 & 0.6 & 0.1 \\ 
SonyAIBORobotSurface1 & 31.7 & nan & 14.3 & 81.7 & 8.8 & 1.8 \\ 
SonyAIBORobotSurface2 & 22.6 & 21.2 & 14.5 & 10.7 & 1.4 & 0.3 \\ 
StarLightCurves & 27.6 & 42.9 & 6.7 & nan & 0.7 & 0.1 \\ 
Strawberry & 70.6 & 20.8 & 333.9 & nan & 0.4 & 0.1 \\ 
SwedishLeaf & 46.3 & 16.5 & 24.7 & 151.5 & 7.9 & 0.4 \\ 
Symbols & 33.8 & 29.9 & 57.2 & 407.6 & 5.6 & 1.6 \\ 
SyntheticControl & 14.7 & 12.6 & 14 & 17.4 & 3.7 & 0.9 \\ 
ToeSegmentation1 & 19.2 & 502.3 & nan & 7 & 4.1 & 0.6 \\ 
ToeSegmentation2 & 3.5 & 2.8 & 154.7 & 89 & 6.2 & 0.3 \\ 
Trace & 93 & 90.8 & 21.7 & 187.4 & 5.7 & 0.7 \\ 
TwoLeadECG & 10.4 & 12.2 & 8.7 & 49.5 & 0.2 & 0.1 \\ 
TwoPatterns & 15.4 & 31.8 & 29.6 & 51.3 & 2.2 & 0.6 \\ 
UMD & 454.1 & 11.4 & 15.5 & 501.3 & 1.2 & 0.4 \\ 
UWaveGestureLibraryAll & 737.2 & nan & nan & 586.1 & 4.5 & 0.2 \\ 
UWaveGestureLibraryX & 31.5 & nan & nan & nan & 6.8 & 0.3 \\ 
UWaveGestureLibraryY & nan & 22 & nan & nan & 7.0 & 0.0 \\ 
UWaveGestureLibraryZ & 9.6 & 24.3 & nan & nan & 27.7 & 2.5 \\ 
Wafer & 26.8 & nan & 23.6 & 164 & 1.5 & 0.3 \\ 
Wine & nan & nan & nan & nan & 0.4 & 0.1 \\ 
WordSynonyms & 34.4 & 11.3 & 42.3 & nan & 2.9 & 0.1 \\ 
Worms & nan & nan & nan & nan & 8.0 & 0.5 \\ 
WormsTwoClass & nan & nan & nan & nan & 8.0 & 1.5 \\ 
Yoga & nan & nan & nan & nan & 2.3 & 0.2 \\ 
\hline
\end{longtable}
\end{center}

\begin{center}
\small
\begin{longtable}{l cccccc}
\caption{Full results of the unconditional sampling experiments with IS.} \\
\label{tab:full_result_is} \\
\hline
Dataset names  & GMMN & RCGAN & TimeGAN & SigCWGAN & TimeVQVAE.mean & TimeVQVAE.std \\
\hline
ACSF1 & 1.4 & 1.9 & 1.5 & 2 & 3.3 & 0.1 \\ 
Adiac & 1.1 & 1.4 & 1.6 & 1.6 & 6.2 & 0.3 \\ 
AllGestureWiimoteX & 3.6 & 3.2 & 4.3 & 1.5 & 2.5 & 0.2 \\ 
AllGestureWiimoteY & 3.4 & 3.2 & 3.1 & 1.7 & 3.0 & 0.1 \\ 
AllGestureWiimoteZ & 2.6 & 2.3 & 2.3 & 1 & 2.6 & 0.0 \\ 
ArrowHead & 1.9 & 1.1 & 1.1 & 1 & 2.5 & 0.1 \\ 
Beef & 1.2 & 1.4 & 1.5 & 1 & 2.7 & 0.2 \\ 
BeetleFly & 1 & 1.1 & 1 & 1 & 1.7 & 0.2 \\ 
BirdChicken & 1 & 1.1 & 1.1 & 1 & 1.8 & 0.0 \\ 
BME & 1 & 1.2 & 1.2 & 1.1 & 2.1 & 0.2 \\ 
Car & 1.7 & 1 & 1 & 1 & 2.3 & 0.1 \\ 
CBF & 1 & 1.1 & 1.5 & 1.1 & 2.7 & 0.0 \\ 
Chinatown & 1.8 & 1 & 1.3 & 1.6 & 2.0 & 0.0 \\ 
ChlorineConcentration & 1.2 & 2 & 1.9 & 1 & 1.6 & 0.1 \\ 
CinCECGTorso & 1.5 & 1.3 & 1.6 & 1 & 1.8 & 0.2 \\ 
Coffee & 1.1 & 1 & 1 & 1 & 1.9 & 0.0 \\ 
Computers & 1.2 & 1.8 & 1.4 & 1.2 & 1.6 & 0.1 \\ 
CricketX & 1.8 & 2 & 2.3 & 1.8 & 3.3 & 0.3 \\ 
CricketY & 2.3 & 3 & 3.6 & 2 & 3.6 & 0.1 \\ 
CricketZ & 3 & 1.9 & 2.5 & 1.9 & 3.3 & 0.1 \\ 
Crop & 5.7 & 7.7 & 7.8 & 7.5 & 17.0 & 0.1 \\ 
DiatomSizeReduction & 1 & 1.2 & 1.3 & 1 & 2.8 & 0.0 \\ 
DistalPhalanxOutlineAgeGroup & 1.5 & 1.6 & 1.3 & 1 & 1.9 & 0.1 \\ 
DistalPhalanxOutlineCorrect & 1.2 & 1.4 & 1.7 & 1.8 & 1.5 & 0.0 \\ 
DistalPhalanxTW & 1.9 & 1.9 & 1.7 & 2.1 & 2.5 & 0.2 \\ 
DodgerLoopDay & 1 & 1.2 & 1.2 & 1.3 & 2.6 & 0.2 \\ 
DodgerLoopGame & 1 & 1.2 & 1 & 1.1 & 1.9 & 0.0 \\ 
DodgerLoopWeekend & 1.2 & 1 & 1.6 & 1.1 & 1.9 & 0.0 \\ 
Earthquakes & 1.2 & 1 & 1 & 1.1 & 1.1 & 0.0 \\ 
ECG200 & 1.4 & 1.5 & 1.4 & 1.5 & 1.5 & 0.0 \\ 
ECG5000 & 1.7 & 1.7 & 1.6 & 1.6 & 2.0 & 0.0 \\ 
ECGFiveDays & 1.1 & 1 & 1.2 & 1.7 & 1.7 & 0.1 \\ 
ElectricDevices & 2.9 & 3.8 & 4.4 & 2.8 & 4.7 & 0.1 \\ 
EOGHorizontalSignal & 1.3 & 2.5 & 1.3 & 2 & 4.7 & 0.2 \\ 
EOGVerticalSignal & 1.5 & 1.7 & 1.7 & 1.2 & 3.2 & 0.1 \\ 
EthanolLevel & 1 & 1.1 & 1 & 2 & 1.3 & 0.1 \\ 
FaceAll & 2 & 2.2 & 3 & 1.8 & 5.3 & 0.2 \\ 
FaceFour & 1.2 & 1.3 & 1.1 & 1.2 & 2.7 & 0.2 \\ 
FacesUCR & 1 & 1.9 & 2.1 & 1.5 & 3.3 & 0.3 \\ 
FiftyWords & 2.7 & 1.9 & 1.3 & 1.8 & 4.6 & 0.2 \\ 
Fish & 1 & 1.2 & 1.2 & 1 & 3.1 & 0.0 \\ 
FordA & 1.5 & 1.3 & 1 & 1 & 1.5 & 0.0 \\ 
FordB & 1 & 1.7 & 1 & 1 & 1.4 & 0.1 \\ 
FreezerRegularTrain & 1 & 1 & 1 & 1.6 & 1.4 & 0.1 \\ 
FreezerSmallTrain & 1.8 & 2 & 1.9 & 1.2 & 1.9 & 0.0 \\ 
Fungi & 1 & 1.1 & 1.3 & 1 & 7.3 & 0.8 \\ 
GestureMidAirD1 & 2.1 & 2.5 & 2.6 & 1.4 & 3.4 & 0.5 \\ 
GestureMidAirD2 & 1.1 & 1.1 & 1.5 & 1 & 4.6 & 0.2 \\ 
GestureMidAirD3 & 3.3 & 2.9 & 1.7 & 2.6 & 2.0 & 0.1 \\ 
GesturePebbleZ1 & 1.2 & 1.1 & 2 & 1.4 & 2.9 & 0.5 \\ 
GesturePebbleZ2 & 1.3 & 1.7 & 1.9 & 1 & 2.8 & 0.2 \\ 
GunPoint & 1.4 & 1.3 & 1.6 & 1 & 1.9 & 0.0 \\ 
GunPointAgeSpan & 1.5 & 1.1 & 1 & 1 & 1.7 & 0.1 \\ 
GunPointMaleVersusFemale & 1 & 1.1 & 1 & 1 & 1.9 & 0.0 \\ 
GunPointOldVersusYoung & 1.7 & 1.3 & 1.9 & 1.3 & 1.9 & 0.0 \\ 
Ham & 1.1 & 1.5 & 1.4 & 1.3 & 1.6 & 0.1 \\ 
HandOutlines & 1.2 & 1 & 1.4 & 1 & 1.2 & 0.0 \\ 
Haptics & 1.9 & 1 & 1 & 1 & 1.9 & 0.0 \\ 
Herring & 1 & 1 & 1.1 & 1 & 1.3 & 0.0 \\ 
HouseTwenty & 1 & 1.2 & 1 & 1 & 1.6 & 0.0 \\ 
InlineSkate & 1.7 & 1.2 & 1.7 & 1 & 1.5 & 0.0 \\ 
InsectEPGRegularTrain & 2.5 & 2.2 & 2.8 & 2.8 & 2.5 & 0.2 \\ 
InsectEPGSmallTrain & 1 & 1 & 1 & 1 & 2.4 & 0.2 \\ 
InsectWingbeatSound & 2.3 & 1.8 & 1.1 & 1.5 & 2.9 & 0.1 \\ 
ItalyPowerDemand & 1.6 & 1.2 & 1.6 & 1.6 & 2.0 & 0.0 \\ 
LargeKitchenAppliances & 1.4 & 1.5 & 1.4 & 1.8 & 2.3 & 0.0 \\ 
Lightning2 & 1.2 & 1.2 & 1.5 & 1 & 1.6 & 0.0 \\ 
Lightning7 & 1 & 1.1 & 1.6 & 1.4 & 3.4 & 0.2 \\ 
Mallat & 1 & 1.1 & 1 & 1 & 4.6 & 0.5 \\ 
Meat & 1.1 & 1.2 & 1.4 & 1 & 1.2 & 0.2 \\ 
MedicalImages & 2.4 & 1.2 & 1.4 & 2.4 & 2.6 & 0.2 \\ 
MelbournePedestrian & 4.2 & 4.8 & 3.8 & 4.7 & 8.9 & 0.1 \\ 
MiddlePhalanxOutlineAgeGroup & 1.6 & 1.6 & 1.7 & 1.1 & 2.0 & 0.1 \\ 
MiddlePhalanxOutlineCorrect & 1.4 & 1.7 & 1.1 & 1.1 & 1.6 & 0.1 \\ 
MiddlePhalanxTW & 2.2 & 2 & 2.3 & 2.4 & 2.8 & 0.1 \\ 
MixedShapesRegularTrain & 2.2 & 1.9 & 2.3 & 1.2 & 1.3 & 0.1 \\ 
MixedShapesSmallTrain & 1.3 & 2.2 & 1.5 & 1 & 2.2 & 0.2 \\ 
MoteStrain & 1.8 & 1.5 & 1.4 & 1.2 & 1.9 & 0.0 \\ 
NonInvasiveFetalECGThorax1 & 2.3 & 1.4 & 2.9 & 1 & 7.6 & 0.7 \\ 
NonInvasiveFetalECGThorax2 & 2 & 1.2 & 2.6 & 1.6 & 7.0 & 1.8 \\ 
OliveOil & 1.2 & 1 & 1 & 1 & 1.0 & 0.0 \\ 
OSULeaf & 2.1 & 1.6 & 1 & 1.1 & 1.4 & 0.0 \\ 
PhalangesOutlinesCorrect & 1.3 & 1.5 & 1.5 & 1 & 1.4 & 0.0 \\ 
Phoneme & 3.7 & 3.7 & 3 & 2 & 2.8 & 0.1 \\ 
PickupGestureWiimoteZ & 1.3 & 1.5 & 1.2 & 1 & 4.7 & 0.3 \\ 
PigAirwayPressure & 2.5 & 1.2 & 1 & 1.2 & 3.8 & 0.7 \\ 
PigArtPressure & 3.6 & 2.9 & 2.1 & 1.2 & 2.1 & 0.2 \\ 
PigCVP & 3 & 1.6 & 1.6 & 1.1 & 3.0 & 0.4 \\ 
PLAID & 1.9 & 2.8 & 2.2 & 1.3 & 1.4 & 0.1 \\ 
Plane & 1.2 & 1.6 & 1.3 & 1 & 5.2 & 0.1 \\ 
PowerCons & 1.2 & 1.1 & 1.1 & 1.5 & 1.7 & 0.0 \\ 
ProximalPhalanxOutlineAgeGroup & 1 & 1.7 & 1.9 & 1.9 & 2.4 & 0.1 \\ 
ProximalPhalanxOutlineCorrect & 1.4 & 1.6 & 1.4 & 1.2 & 1.6 & 0.1 \\ 
ProximalPhalanxTW & 1.5 & 2.2 & 1.8 & 1.1 & 2.7 & 0.1 \\ 
RefrigerationDevices & 1.3 & 1.2 & 2 & 1.3 & 1.6 & 0.2 \\ 
Rock & 1 & 1 & 1 & 1 & 2.5 & 0.2 \\ 
ScreenType & 1.4 & 1.2 & 1.2 & 1.3 & 2.2 & 0.0 \\ 
SemgHandGenderCh2 & 1 & 1.3 & 1 & 1 & 1.4 & 0.0 \\ 
SemgHandMovementCh2 & 2.2 & 1.8 & 1.4 & 1.1 & 1.9 & 0.1 \\ 
SemgHandSubjectCh2 & 2 & 3.6 & 3.1 & 1.1 & 2.0 & 0.1 \\ 
ShakeGestureWiimoteZ & 1.9 & 2.5 & 2.4 & 1.3 & 5.3 & 0.2 \\ 
ShapeletSim & 1.1 & 1.1 & 1.2 & 1.1 & 1.8 & 0.1 \\ 
ShapesAll & 1.4 & 3.2 & 2.7 & 1.3 & 5.9 & 0.1 \\ 
SmallKitchenAppliances & 1.3 & 1.4 & 1.2 & 1.6 & 1.6 & 0.0 \\ 
SmoothSubspace & 1 & 1.9 & 1.9 & 1.8 & 2.6 & 0.0 \\ 
SonyAIBORobotSurface1 & 1.1 & 1.1 & 1.3 & 1 & 1.7 & 0.1 \\ 
SonyAIBORobotSurface2 & 1.3 & 1.2 & 1.5 & 1.5 & 1.9 & 0.0 \\ 
StarLightCurves & 1.8 & 1.8 & 1.6 & 1.1 & 2.3 & 0.1 \\ 
Strawberry & 1.3 & 1 & 1 & 1 & 1.8 & 0.0 \\ 
SwedishLeaf & 1.6 & 2.5 & 2 & 1.6 & 7.2 & 0.1 \\ 
Symbols & 1.3 & 1.1 & 2.2 & 1.3 & 4.0 & 0.2 \\ 
SyntheticControl & 2.4 & 2.3 & 2.4 & 2.1 & 4.4 & 0.2 \\ 
ToeSegmentation1 & 1 & 1.4 & 1.7 & 1.3 & 1.7 & 0.1 \\ 
ToeSegmentation2 & 1.2 & 1.2 & 1.6 & 1 & 1.8 & 0.0 \\ 
Trace & 1.1 & 1 & 1.5 & 1 & 3.0 & 0.1 \\ 
TwoLeadECG & 1.3 & 1.3 & 1.4 & 1.2 & 1.9 & 0.0 \\ 
TwoPatterns & 1.6 & 1.7 & 1.6 & 1.6 & 2.8 & 0.1 \\ 
UMD & 1 & 1.8 & 1 & 1.3 & 2.5 & 0.0 \\ 
UWaveGestureLibraryAll & 2.4 & 2.1 & 3.2 & 1.9 & 3.2 & 0.1 \\ 
UWaveGestureLibraryX & 2.3 & 2.2 & 1 & 2.2 & 3.9 & 0.0 \\ 
UWaveGestureLibraryY & 3.2 & 1.8 & 1 & 1.8 & 3.3 & 0.1 \\ 
UWaveGestureLibraryZ & 2.6 & 2.9 & 2.2 & 1.1 & 3.0 & 0.1 \\ 
Wafer & 1.2 & nan & 1.1 & 1.6 & 1.3 & 0.0 \\ 
Wine & 1.1 & 1.8 & 1.1 & 1 & 1.7 & 0.1 \\ 
WordSynonyms & 1.3 & 1.6 & 1.4 & 1.4 & 1.9 & 0.0 \\ 
Worms & 1.2 & 1.5 & 1.1 & 1.1 & 2.0 & 0.1 \\ 
WormsTwoClass & 1.2 & 1.8 & 1.5 & 1.5 & 1.0 & 0.0 \\ 
Yoga & 1.5 & 1.1 & 1 & 1 & 1.4 & 0.0 \\ 
\hline
\end{longtable}
\end{center}

\begin{center}
\small
\begin{longtable}{l cccccc}
\caption{Full results of the class-conditional sampling experiments with CAS.} \\
\label{tab:full_result_cas} \\
\hline
Dataset names  & WGAN \citep{smith2020one} & TSGAN \citep{smith2020one} & TimeVQVAE.mean & TimeVQVAE.std \\
\hline
ACSF1 & ~ & ~ & 72.7 & 3.2 \\ 
Adiac & ~ & ~ & 77.2 & 2.5 \\ 
AllGestureWiimoteX & ~ & ~ & 55.6 & 1.6 \\ 
AllGestureWiimoteY & ~ & ~ & 67.3 & 1.0 \\ 
AllGestureWiimoteZ & ~ & ~ & 62.2 & 2.3 \\ 
ArrowHead & 61.7 & 85.7 & 81.7 & 4.0 \\ 
BME & 80 & 82.7 & 75.3 & 1.2 \\ 
Beef & 20 & 60 & 72.2 & 8.4 \\ 
BeetleFly & 55 & 90 & 91.7 & 5.8 \\ 
BirdChicken & 90 & 75 & 83.3 & 7.6 \\ 
CBF & 70.9 & 87.3 & 96.5 & 0.4 \\ 
Car & 65 & 70 & 83.9 & 13.6 \\ 
Chinatown & ~ & ~ & 97.8 & 0.7 \\ 
ChlorineConcentration & 56.5 & 54.5 & 70.5 & 0.8 \\ 
CinCECGTorso & 40.6 & 49.1 & 75.5 & 5.5 \\ 
Coffee & 100 & 100 & 100.0 & 0.0 \\ 
Computers & 50.8 & 65.2 & 67.1 & 3.2 \\ 
CricketX & ~ & ~ & 70.3 & 3.0 \\ 
CricketY & ~ & ~ & 71.7 & 2.6 \\ 
CricketZ & ~ & ~ & 71.7 & 1.6 \\ 
Crop & ~ & ~ & 67.6 & 0.7 \\ 
DiatomSizeReduction & 73.5 & 79.7 & 92.3 & 2.3 \\ 
DistalPhalanxOutlineAgeGroup & 73.4 & 70.5 & 73.9 & 5.1 \\ 
DistalPhalanxOutlineCorrect & 68.8 & 64.9 & 78.7 & 1.3 \\ 
DistalPhalanxTW & ~ & ~ & 69.3 & 2.5 \\ 
DodgerLoopDay & ~ & ~ & 47.5 & 4.5 \\ 
DodgerLoopGame & ~ & ~ & 69.1 & 3.0 \\ 
DodgerLoopWeekend & ~ & ~ & 97.1 & 0.7 \\ 
ECG200 & 82 & 82 & 89.0 & 1.0 \\ 
ECG5000 & 78.2 & 86 & 93.3 & 0.2 \\ 
ECGFiveDays & 92.6 & 98.8 & 96.3 & 2.6 \\ 
EOGHorizontalSignal & ~ & ~ & 58.6 & 1.5 \\ 
EOGVerticalSignal & ~ & ~ & 47.3 & 0.8 \\ 
Earthquakes & 69.1 & 74.8 & 73.4 & 0.7 \\ 
ElectricDevices & ~ & ~ & 63.5 & 0.2 \\ 
EthanolLevel & 24.8 & 30.2 & 43.7 & 6.3 \\ 
FaceAll & ~ & ~ & 83.2 & 0.8 \\ 
FaceFour & 65.2 & 84.2 & 92.4 & 3.3 \\ 
FacesUCR & ~ & ~ & 93.2 & 0.4 \\ 
FiftyWords & ~ & ~ & 62.8 & 0.6 \\ 
Fish & ~ & ~ & 92.2 & 0.3 \\ 
FordA & 80 & 89.2 & 87.2 & 3.7 \\ 
FordB & 62.2 & 61.4 & 68.7 & 8.9 \\ 
FreezerRegularTrain & 50 & 50.1 & 93.9 & 5.2 \\ 
FreezerSmallTrain & 50.7 & 76 & 77.4 & 1.9 \\ 
Fungi & ~ & ~ & 98.6 & 0.3 \\ 
GestureMidAirD1 & ~ & ~ & 67.2 & 2.5 \\ 
GestureMidAirD2 & ~ & ~ & 54.4 & 0.9 \\ 
GestureMidAirD3 & ~ & ~ & 26.7 & 2.4 \\ 
GesturePebbleZ1 & ~ & ~ & 82.6 & 2.5 \\ 
GesturePebbleZ2 & ~ & ~ & 78.5 & 2.2 \\ 
GunPoint & 100 & 100 & 99.1 & 1.0 \\ 
GunPointAgeSpan & 41.3 & 44.1 & 98.5 & 0.3 \\ 
GunPointMaleVersusFemale & 52.5 & 52.5 & 99.9 & 0.1 \\ 
GunPointOldVersusYoung & 11.4 & 52.4 & 100.0 & 0.0 \\ 
Ham & 68.6 & 68.6 & 70.2 & 2.7 \\ 
HandOutlines & 64.1 & 65.7 & 75.7 & 9.8 \\ 
Haptics & 20.8 & 31.5 & 44.7 & 1.9 \\ 
Herring & 46.9 & 65.6 & 59.4 & 0.0 \\ 
HouseTwenty & 58 & 42.1 & 88.0 & 5.7 \\ 
InlineSkate & ~ & ~ & 28.7 & 4.2 \\ 
InsectEPGRegularTrain & 64.3 & 64.3 & 100.0 & 0.0 \\ 
InsectEPGSmallTrain & 16.9 & 35.7 & 100.0 & 0.0 \\ 
InsectWingbeatSound & ~ & ~ & 38.0 & 1.9 \\ 
ItalyPowerDemand & ~ & ~ & 95.3 & 1.2 \\ 
LargeKitchenAppliances & 59.5 & 74.9 & 84.1 & 1.0 \\ 
Lightning2 & 70.5 & 72.2 & 74.3 & 7.4 \\ 
Lightning7 & ~ & ~ & 69.4 & 3.2 \\ 
Mallat & ~ & ~ & 96.4 & 0.6 \\ 
Meat & 88.3 & 46.7 & 68.9 & 17.7 \\ 
MedicalImages & ~ & ~ & 73.2 & 2.6 \\ 
MelbournePedestrian & ~ & ~ & 94.2 & 0.3 \\ 
MiddlePhalanxOutlineAgeGroup & 40.9 & 50 & 54.8 & 2.5 \\ 
MiddlePhalanxOutlineCorrect & 73.5 & 75.3 & 78.5 & 3.3 \\ 
MiddlePhalanxTW & ~ & ~ & 49.6 & 3.1 \\ 
MixedShapesRegularTrain & ~ & ~ & 79.2 & 4.3 \\ 
MixedShapesSmallTrain & 55.5 & 82.6 & 82.7 & 6.8 \\ 
MoteStrain & 87.9 & 87.3 & 90.1 & 0.6 \\ 
NonInvasiveFetalECGThorax1 & ~ & ~ & 72.6 & 5.8 \\ 
NonInvasiveFetalECGThorax2 & ~ & ~ & 75.4 & 7.9 \\ 
OSULeaf & ~ & ~ & 83.5 & 1.1 \\ 
OliveOil & 40 & 40 & 31.1 & 15.4 \\ 
PLAID & ~ & ~ & 35.8 & 2.2 \\ 
PhalangesOutlinesCorrect & 73.9 & 76.9 & 79.0 & 1.0 \\ 
Phoneme & ~ & ~ & 28.0 & 1.0 \\ 
PickupGestureWiimoteZ & ~ & ~ & 75.3 & 3.1 \\ 
PigAirwayPressure & ~ & ~ & 39.1 & 10.9 \\ 
PigArtPressure & ~ & ~ & 97.0 & 1.0 \\ 
PigCVP & ~ & ~ & 27.7 & 6.5 \\ 
Plane & ~ & ~ & 100.0 & 0.0 \\ 
PowerCons & 51.1 & 52.2 & 89.4 & 0.6 \\ 
ProximalPhalanxOutlineAgeGroup & 87.3 & 85.9 & 81.5 & 2.2 \\ 
ProximalPhalanxOutlineCorrect & 84.5 & 88.3 & 88.8 & 4.3 \\ 
ProximalPhalanxTW & ~ & ~ & 75.6 & 2.0 \\ 
RefrigerationDevices & 33.1 & 42.4 & 40.4 & 1.4 \\ 
Rock & 20 & 36 & 58.0 & 6.9 \\ 
ScreenType & 52 & 56.8 & 54.3 & 3.5 \\ 
SemgHandGenderCh2 & 65.2 & 65.2 & 60.5 & 9.6 \\ 
SemgHandMovementCh2 & ~ & ~ & 31.5 & 2.6 \\ 
SemgHandSubjectCh2 & 26 & 29.3 & 28.8 & 3.7 \\ 
ShakeGestureWiimoteZ & ~ & ~ & 92.0 & 3.5 \\ 
ShapeletSim & 50 & 50 & 82.8 & 14.5 \\ 
ShapesAll & ~ & ~ & 74.5 & 2.7 \\ 
SmallKitchenAppliances & 56.5 & 64 & 72.4 & 5.9 \\ 
SmoothSubspace & 66.7 & 68 & 96.2 & 2.7 \\ 
SonyAIBORobotSurface1 & 89.5 & 92.8 & 98.1 & 0.6 \\ 
SonyAIBORobotSurface2 & 92.2 & 84.7 & 97.2 & 0.6 \\ 
StarLightCurves & 48.7 & 81.4 & 95.6 & 1.1 \\ 
Strawberry & 83.2 & 93.2 & 95.7 & 0.8 \\ 
SwedishLeaf & ~ & ~ & 95.0 & 0.3 \\ 
Symbols & ~ & ~ & 96.8 & 0.7 \\ 
SyntheticControl & ~ & ~ & 97.2 & 0.8 \\ 
ToeSegmentation1 & 88.6 & 93.4 & 92.0 & 1.8 \\ 
ToeSegmentation2 & 80.8 & 91.5 & 86.9 & 2.3 \\ 
Trace & 97 & 100 & 97.7 & 2.5 \\ 
TwoLeadECG & 99.5 & 99.8 & 97.6 & 1.3 \\ 
TwoPatterns & 76.7 & 86.8 & 86.5 & 0.5 \\ 
UMD & 97.2 & 97.9 & 99.3 & 0.0 \\ 
UWaveGestureLibraryAll & ~ & ~ & 59.4 & 3.3 \\ 
UWaveGestureLibraryX & ~ & ~ & 63.7 & 0.6 \\ 
UWaveGestureLibraryY & ~ & ~ & 56.1 & 1.4 \\ 
UWaveGestureLibraryZ & ~ & ~ & 60.8 & 4.3 \\ 
Wafer & 91.5 & 79.2 & 99.4 & 0.1 \\ 
Wine & 55.6 & 61.1 & 61.7 & 8.4 \\ 
WordSynonyms & ~ & ~ & 52.4 & 1.8 \\ 
Worms & 44.2 & 59.7 & 41.1 & 0.7 \\ 
WormsTwoClass & 71.4 & 76.6 & 63.2 & 2.0 \\ 
Yoga & 53.6 & 84.5 & 74.1 & 1.8 \\ 
\hline
\end{longtable}
\end{center}

\subsection{Quantitative Result Summaries of the Ablation Studies}
\begin{figure}[ht!]
\includegraphics[width=\linewidth]{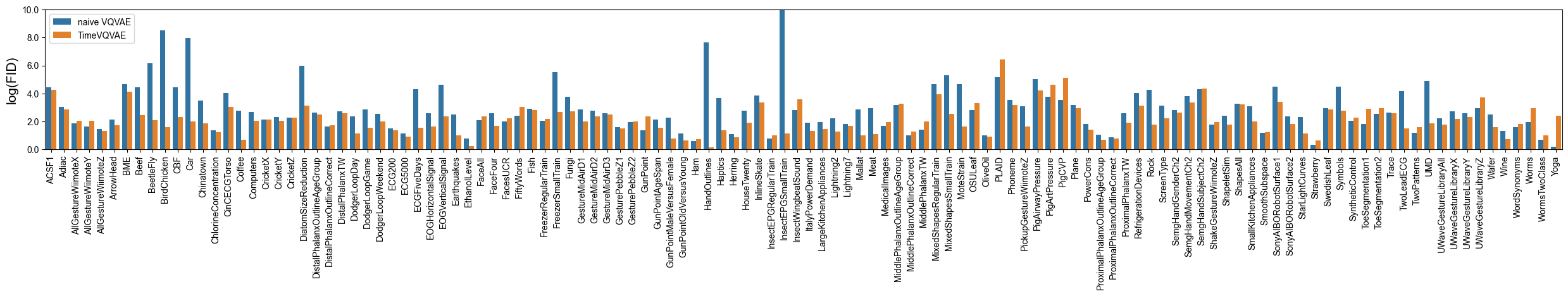}
\includegraphics[width=\linewidth]{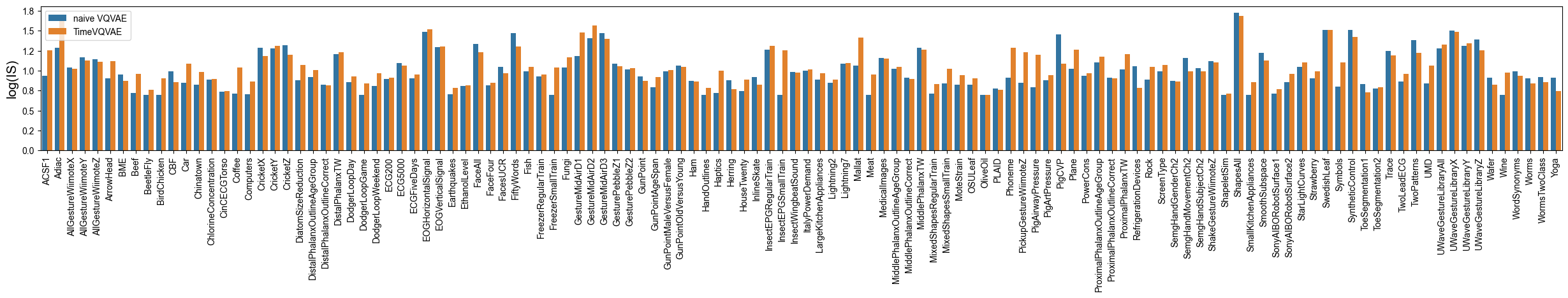}
\caption{Quantitative result summary of the following ablation study: \textbf{Naive VQ-VAE vs TimeVQVAE}} 
\label{fig:bargraph_naiveVQVAE}
\end{figure}

\begin{figure}[ht!]
\includegraphics[width=\linewidth]{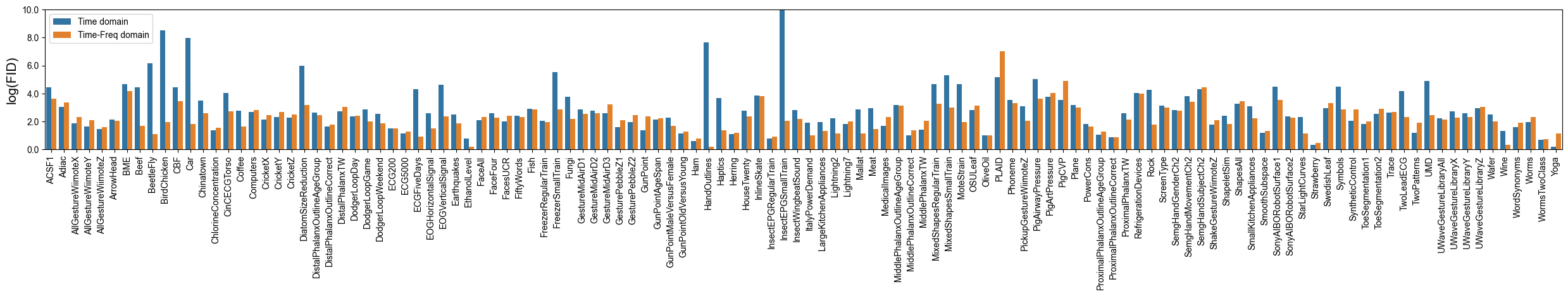}
\includegraphics[width=\linewidth]{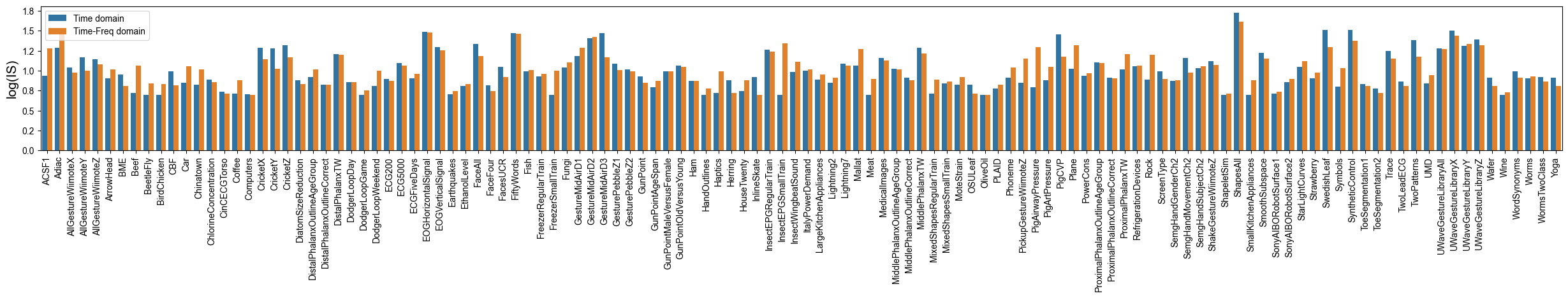}
\caption{Quantitative result summary of the following ablation study: \textbf{VQ in Time and Time-frequency Domains}} 
\label{fig:bargraph_time_vs_timefreqE}
\end{figure}

\begin{figure}[ht!]
\includegraphics[width=\linewidth]{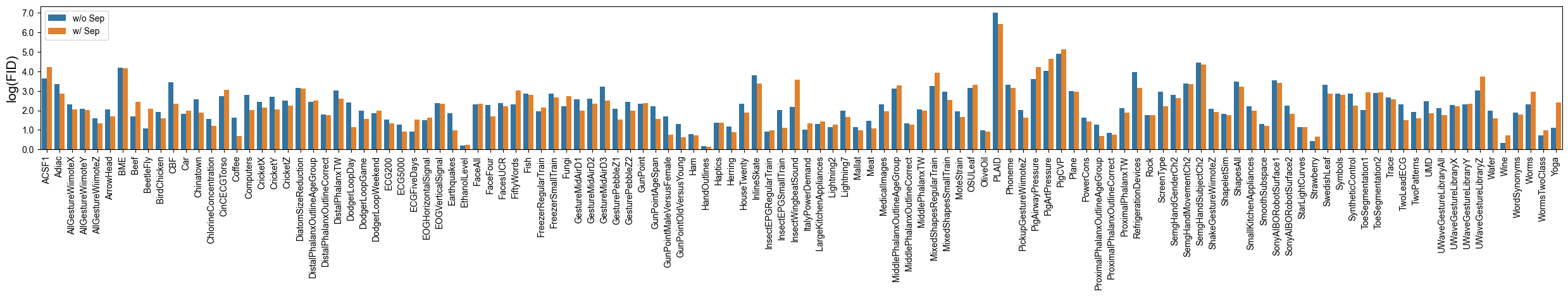}
\includegraphics[width=\linewidth]{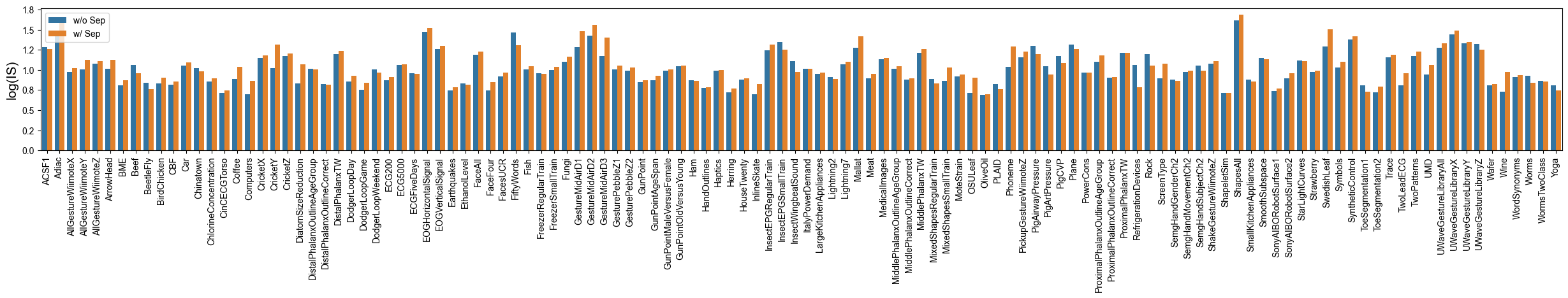}
\caption{Quantitative result summary of the following ablation study: \textbf{Separation of LF and HF Latent Spaces}} 
\label{fig:bargraph_LF_HF_sep}
\end{figure}

\begin{figure}[ht!]
\includegraphics[width=\linewidth]{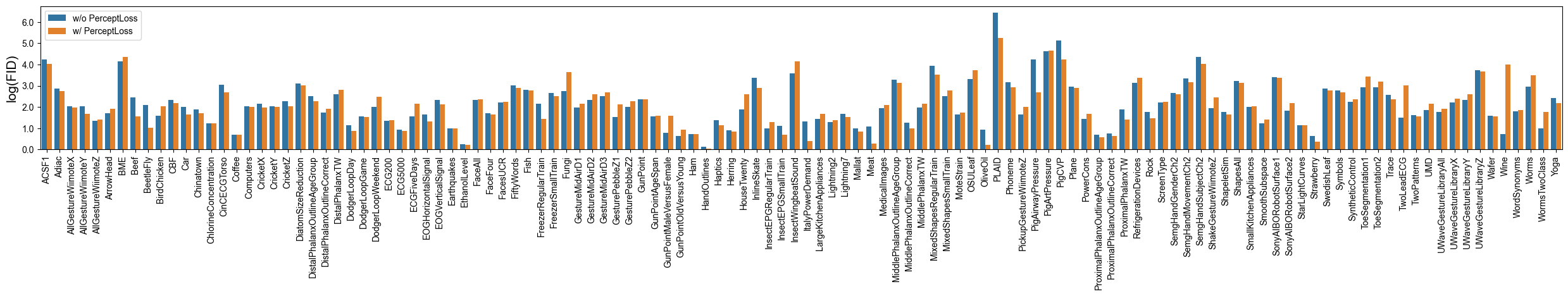}
\includegraphics[width=\linewidth]{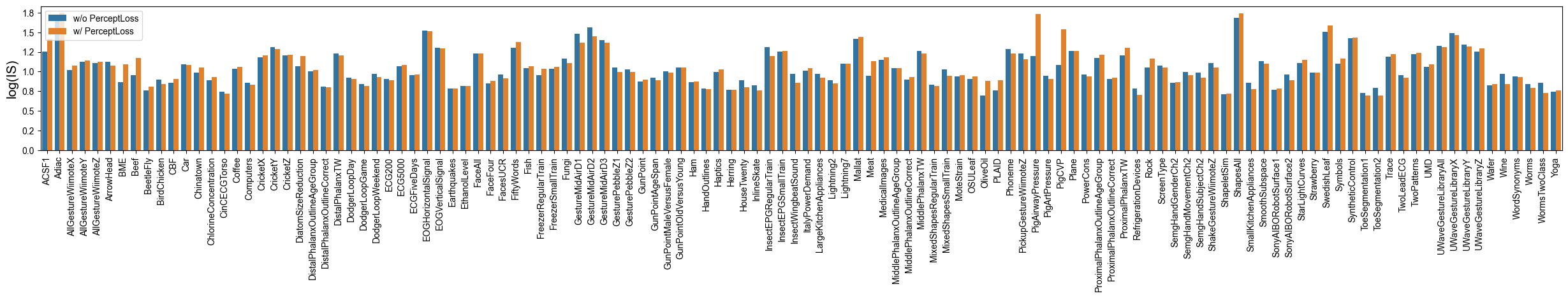}
\caption{Quantitative result summary of the following ablation study: \textbf{Perceptual Loss}} 
\label{fig:bargraph_perceptual_loss}
\end{figure}

The ablation study results in the main paper are presented by critical diagrams. Here we present those results more in detail. Figs. \ref{fig:bargraph_naiveVQVAE}-\ref{fig:bargraph_perceptual_loss} present bar graphs where the x-axis represents dataset names and y-axis represents a metric such as FID or IS. For FID, the lower score denotes better performance. For IS, the higher score denotes better performance.

\section{ADDITIONAL ABLATION STUDY}



\subsection{The Independence Assumptions}
\begin{figure*}[ht!]
\centering
\includegraphics[width=0.95\linewidth]{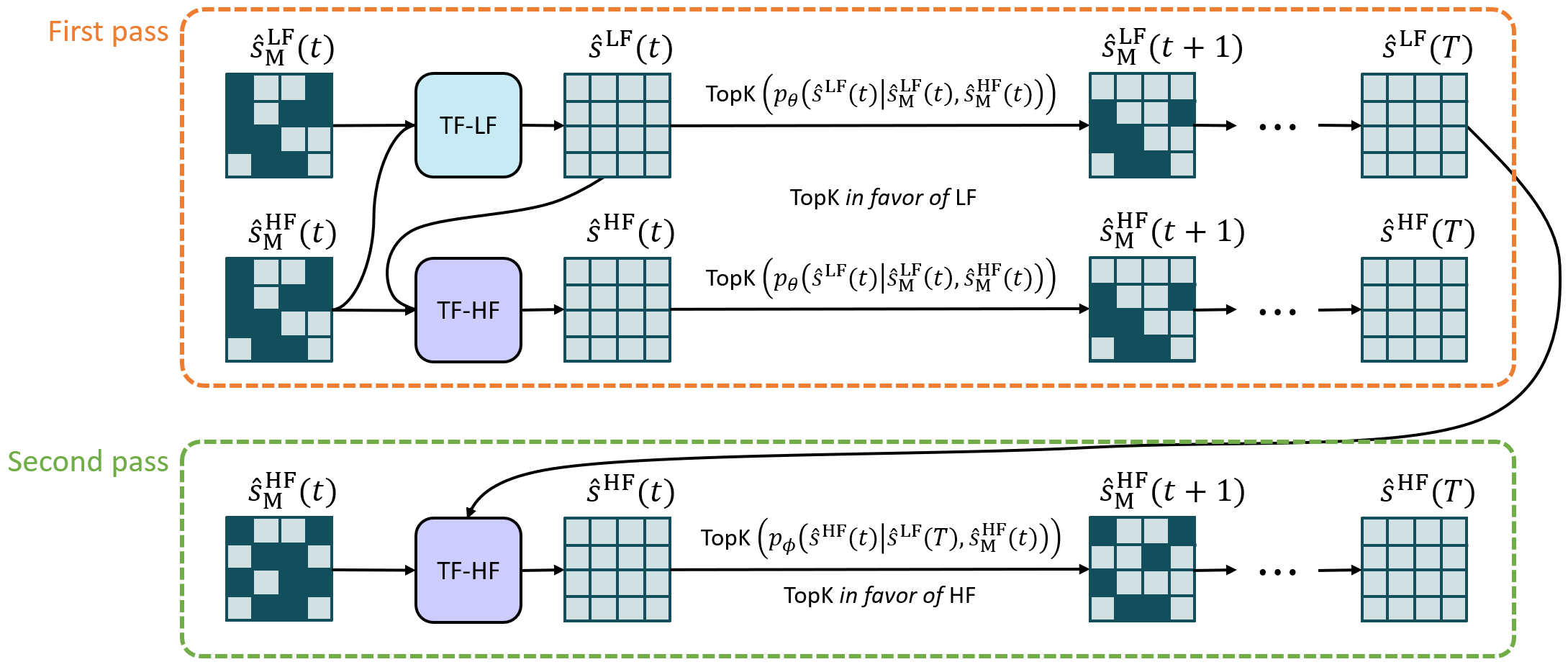}
\caption{Overview of the iterative decoding without the independence assumptions. It is designed to maximize both $p(\hat{s}^{\mathrm{LF}} | \cdot)$ and $p(\hat{s}^{\mathrm{HF}} | \cdot)$. In the first pass, $\mathrm{TopK}(p_\theta(\hat{s}^\mathrm{LF}(t)|\cdot))$ operations are used for both LF and HF. That is because we need to have the same masking regions for both $\hat{s}^\mathrm{LF}(t)$ and $\hat{s}^\mathrm{HF}(t)$ according to $p(s_\mathrm{M}^\mathrm{LF}, s_\mathrm{M}^\mathrm{HF})$. One could choose $\mathrm{TopK}(p_\theta(\hat{s}^\mathrm{LF}(t)|\cdot) p_\theta(\hat{s}^\mathrm{HF}(t)|\cdot))$, but we experimentally found that it results in poor quality LF sampling.}
\label{fig:two_staged_maskgit_wo_assumption}
\end{figure*}

\begin{figure}[h]
\includegraphics[width=0.5\linewidth]{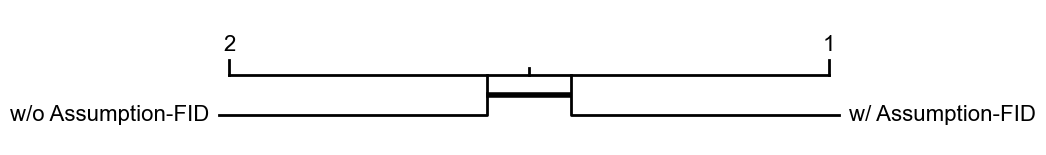}
\includegraphics[width=0.5\linewidth]{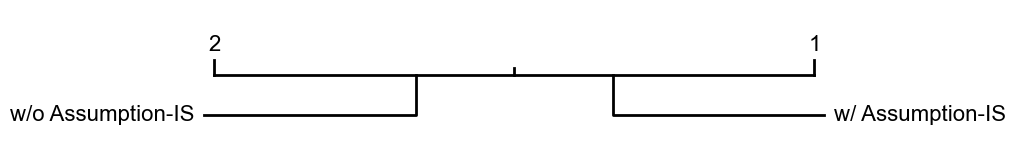}
\includegraphics[width=\linewidth]{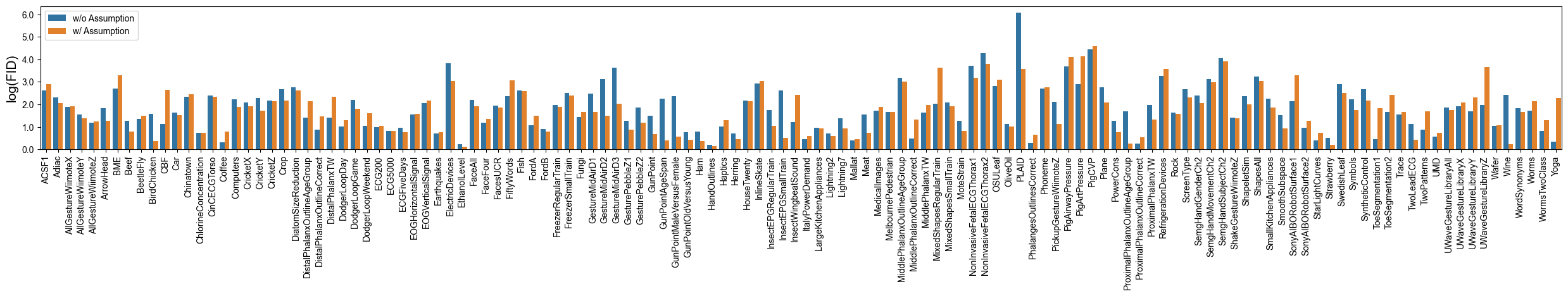}
\includegraphics[width=\linewidth]{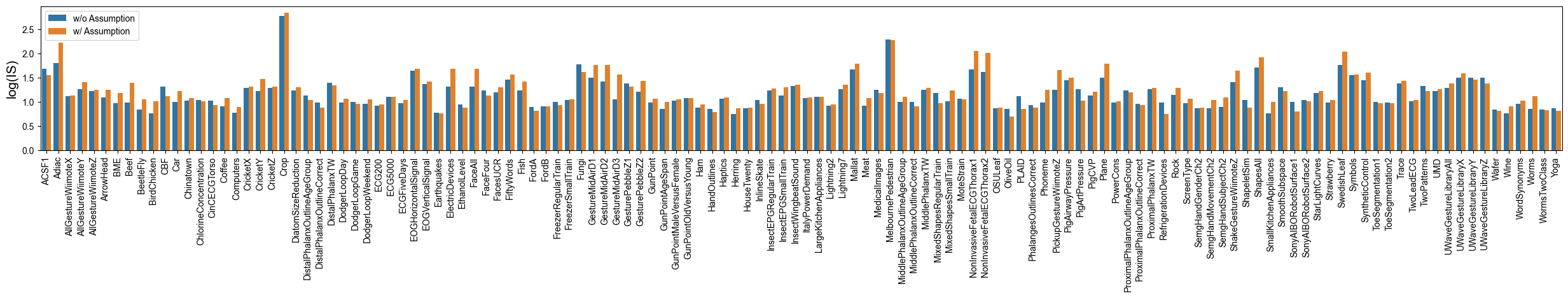}
\caption{Performance comparison with respect to the proposed iterative decoding with (w/) and without (w/o) the independence assumptions.}
\label{fig:comp_wrt_iterative_decoding_w_wo_assumption}
\end{figure}

We simplify Eq.~\eqref{eq:p_s} with the following independence assumptions -- $p(s^{\mathrm{LF}} | s_\mathrm{M}^{\mathrm{HF}}) = p(s^{\mathrm{LF}})$ and $p(s_\mathrm{M}^\mathrm{LF}, s_\mathrm{M}^\mathrm{HF}) = p(s_\mathrm{M}^\mathrm{LF}) p(s_\mathrm{M}^\mathrm{HF})$. Such simplification greatly reduces complexity of the proposed iterative decoding process, which would look like Fig.~\ref{fig:two_staged_maskgit_wo_assumption}, if not assuming independence in Eq.~\eqref{eq:p_s}. Without the independence assumptions, the sampling of $\hat{s}^\mathrm{LF}$ is tangled with that of $\hat{s}^\mathrm{HF}$, therefore, the downsampling rate of the LF and HF encoders must be the same.
The performance comparison between the iterative decoding using Eq. \eqref{eq:p_s} and Eq. \eqref{eq:p_s_simplified} is shown in Fig. \ref{fig:comp_wrt_iterative_decoding_w_wo_assumption}. 
In this experiment, the model without the assumptions uses the downsampled width of 32 for both LF and HF. Both cases use the \textit{Small}-sized encoder, decoder, and prior model. The performance with the assumptions is generally better. That is due to the higher downsampling rate for LF which enables to capture the global consistency better. The model without the assumptions could use the downsampled width of 8 to improve the global consistency but that, on the other side, would cause  the loss of  HF reconstruction details.

\subsection{Limitation of the Naive VQ-VAE for Time Series}
\begin{figure}[ht!]
\centering
\includegraphics[width=0.8\linewidth]{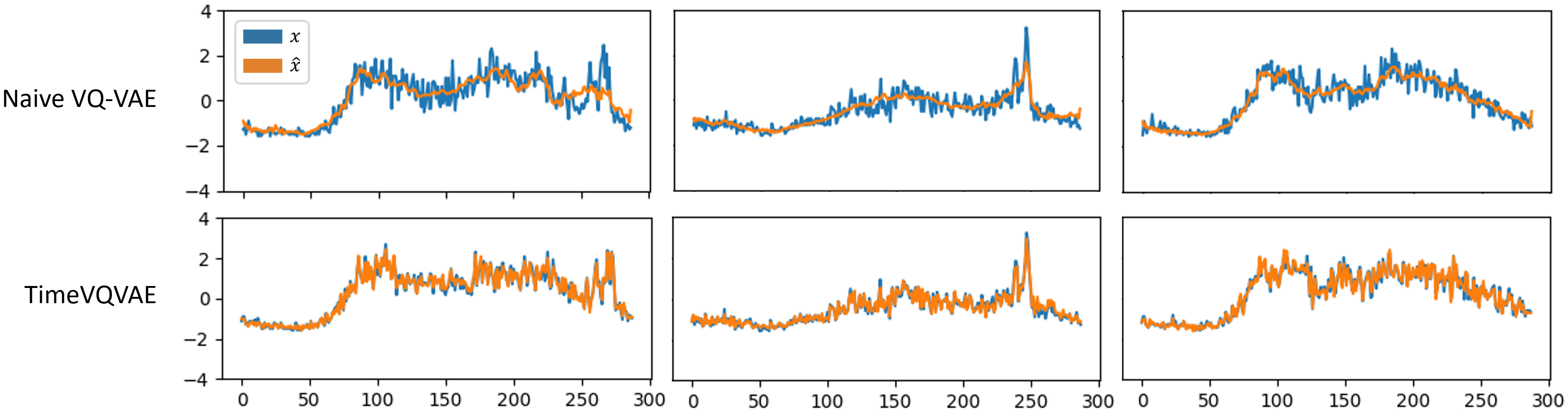}
\caption{Reconstruction examples by the naive VQ-VAE and TimeVQVAE.}
\label{fig:reconstruction_comp}
\end{figure}

\begin{figure}[ht!]
\centering
\includegraphics[width=0.75\linewidth]{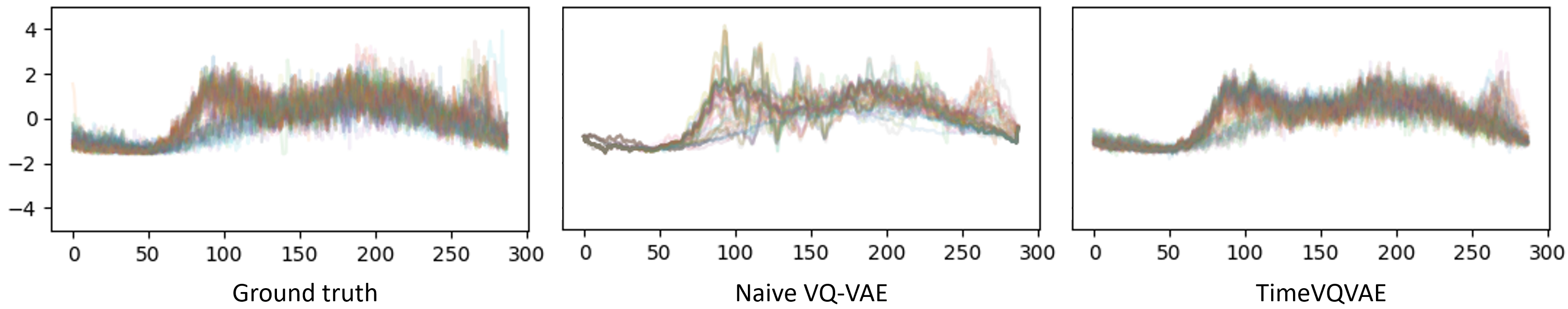}
\caption{Visualization of generated samples by the naive VQ-VAE and TimeVQVAE.}
\label{fig:reconstruction_comp_generated_samples}
\end{figure}

We have experimentally found that the naive form VQ-VAE for time series has difficulty with reconstructing time series, especially HF components. TimeVQVAE overcomes the limitation with the LF-HF separation where two sets of an encoder and decoder are dedicated to different frequency bands -- one for LF and another for HF. 
Fig.~\ref{fig:reconstruction_comp} presents examples of reconstruction by the naive VQ-VAE and TimeVQVAE, and shows that the naive VQ-VAE struggles with reconstructing the HF components. As a result, the naive VQ-VAE results in poorer-quality generated samples. Fig.~\ref{fig:reconstruction_comp_generated_samples} shows generated samples by the naive VQ-VAE and TimeVQVAE on a dataset, DodgerLoopDay. The poor VQ modeling in the naive VQ-VAE results in poor quality of the synthetic samples. The same phenomenon can be observed for datasets such as Computers and Earthquakes.

\subsection{Guided Class-Conditional Sampling}
\begin{figure}[ht!]
\centering
\includegraphics[width=\linewidth]{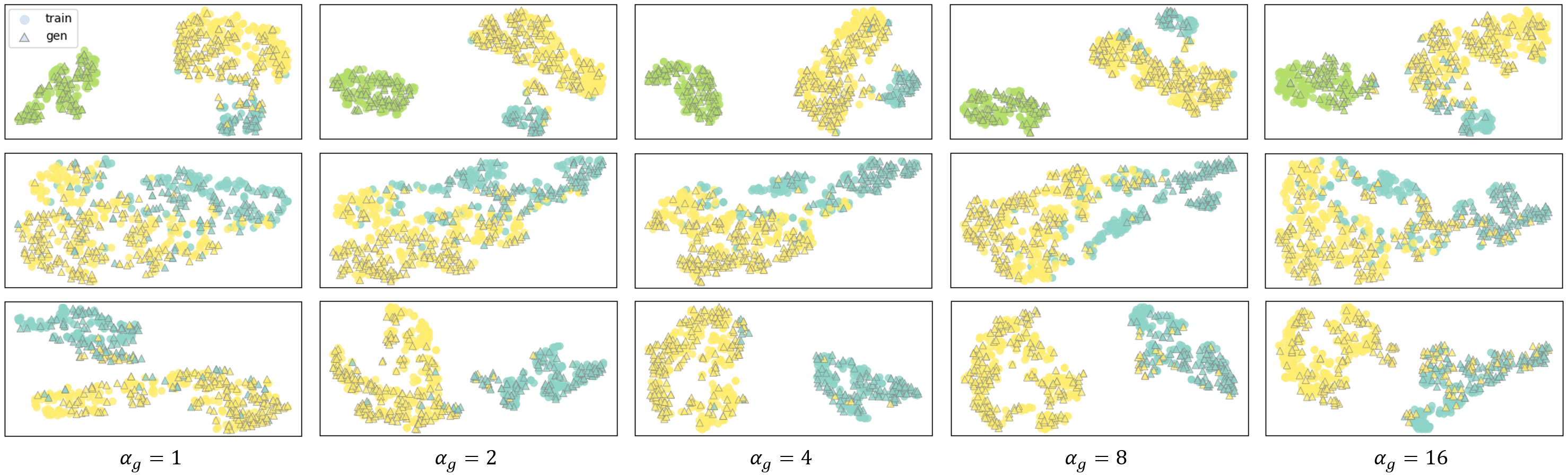}
\caption{2-dimensional t-SNE mapping of representations of real samples (train) and generated samples (gen). The representations are obtained with the pretrained FCN. The different colors denote different classes. The different rows represent different datasets. The used datasets are StarLightCurves, HandOutlines, and MiddlePhalanxOutlineCorrect, in order.}
\label{fig:guided_scale_ablation}
\end{figure}

The guidance scale, $\alpha_g$ generally improves quality of generated samples to be more coherent with their class information. $\alpha_g$, however, cannot be indefinitely increased for the further class-coherence. \cite{saharia2022photorealistic} found that a train-test mismatch arises from a high guidance scale, damaging fidelity of generated samples. We have discovered the same phenomenon. 
Fig.~\ref{fig:guided_scale_ablation} shows the effects of the different guidance scales on multiple datasets. The class boundaries of the generated samples improve as $\alpha_g$ increases until around 4, but soon start deteriorating as the scale gets significantly larger due to the train-test mismatch.

\subsection{Different Model Size Parameters}
\begin{figure*}[ht!]
\includegraphics[width=\linewidth]{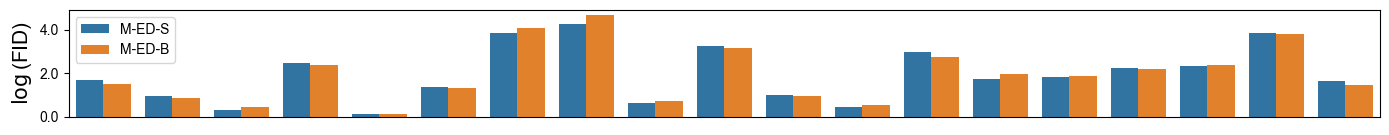}
\includegraphics[width=\linewidth]{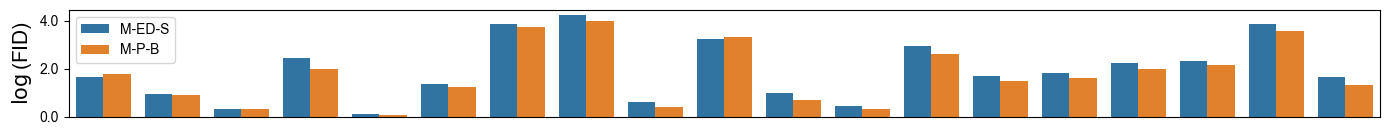}
\includegraphics[width=\linewidth]{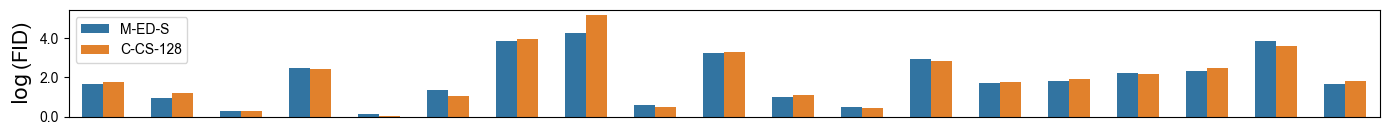}
\includegraphics[width=\linewidth]{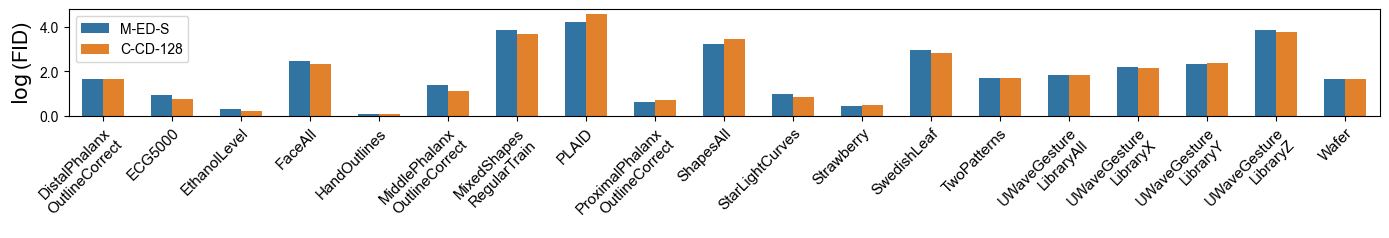}
\caption{Ablation study results with respect to the model size of TimeVQVAE. The y-axis represents a log-scaled FID score. Specifically, it uses $\log(\mathrm{FID} + 1.)$ to prevent a negative value. The associated IS shows similar ranking orders as the FID. They are omitted for brevity.} 
\label{fig:timevqvae_model_param_ablation}
\end{figure*}

Table \ref{tab:case_ind_timevqvae_ablation_study} introduces different experimental cases for the model-size ablation study. The sizes -- Small and Base -- for the encoder, decoder, and prior model are detailed in Tables \ref{tab:size_enc_dec}-\ref{tab:size_prior_model}. 
For this ablation study, datasets with training set sizes between 500 and 1,000 are used. The optimizer settings are the same as described in Sect. \ref{sect:optimizer} in this subsection except that maximum epochs of \{stage~1: 1,000, stage~2: 5,000\} are used.
Fig. \ref{fig:timevqvae_model_param_ablation} shows the ablation study results. The larger encoder, decoder, and prior model seem to have a positive effect. Among them, the larger prior model results in a clear performance gain. The codebook size and dimension, on the other hand, do not show a clear impact.

\begin{table*}[ht!]
    \caption{Case indices for the model size ablation study for TimeVQVAE.} 
    \label{tab:case_ind_timevqvae_ablation_study}
    \centering
    \begin{tabular}{l| c c c | c c |l}
    \hline
        Ablation on & \rot{\shortstack[l]{Encoder\\size}} & \rot{\shortstack[l]{Decoder\\size}} & \rot{\shortstack[l]{Prior model\: \\size}} &  \rot{\shortstack[l]{Codebook \\size}} & \rot{Code dim}  & Case index \\ \hline
        \multirow{5}{*}{Model size} & Small & Small & \textcolor{lightgray}{Small} & \textcolor{lightgray}{32} & \textcolor{lightgray}{32} & M-ED-S \\ 
                                    & Base & Base & \textcolor{lightgray}{Small} & \textcolor{lightgray}{32} & \textcolor{lightgray}{32}  & M-ED-B \\
                                    & \textcolor{lightgray}{Small} & \textcolor{lightgray}{Small} & Base & \textcolor{lightgray}{32} & \textcolor{lightgray}{32}  & M-P-B \\ \hline
        \multirow{3}{*}{Codebook learning} & \textcolor{lightgray}{Small} & \textcolor{lightgray}{Small} & \textcolor{lightgray}{Small} & 128 & \textcolor{lightgray}{32} & C-CS-128 \\
                                           & \textcolor{lightgray}{Small} & \textcolor{lightgray}{Small} & \textcolor{lightgray}{Small} & \textcolor{lightgray}{32} & 128 & C-CD-128 \\ \hline
    \end{tabular}
\end{table*}


\section{LIMITATIONS OF THE EXISTING EVALUATION METRICS IN TSG}
\label{appendix: limitation of the existing evaluation metrics in tsg}

\begin{figure}[h!]
\includegraphics[width=\linewidth]{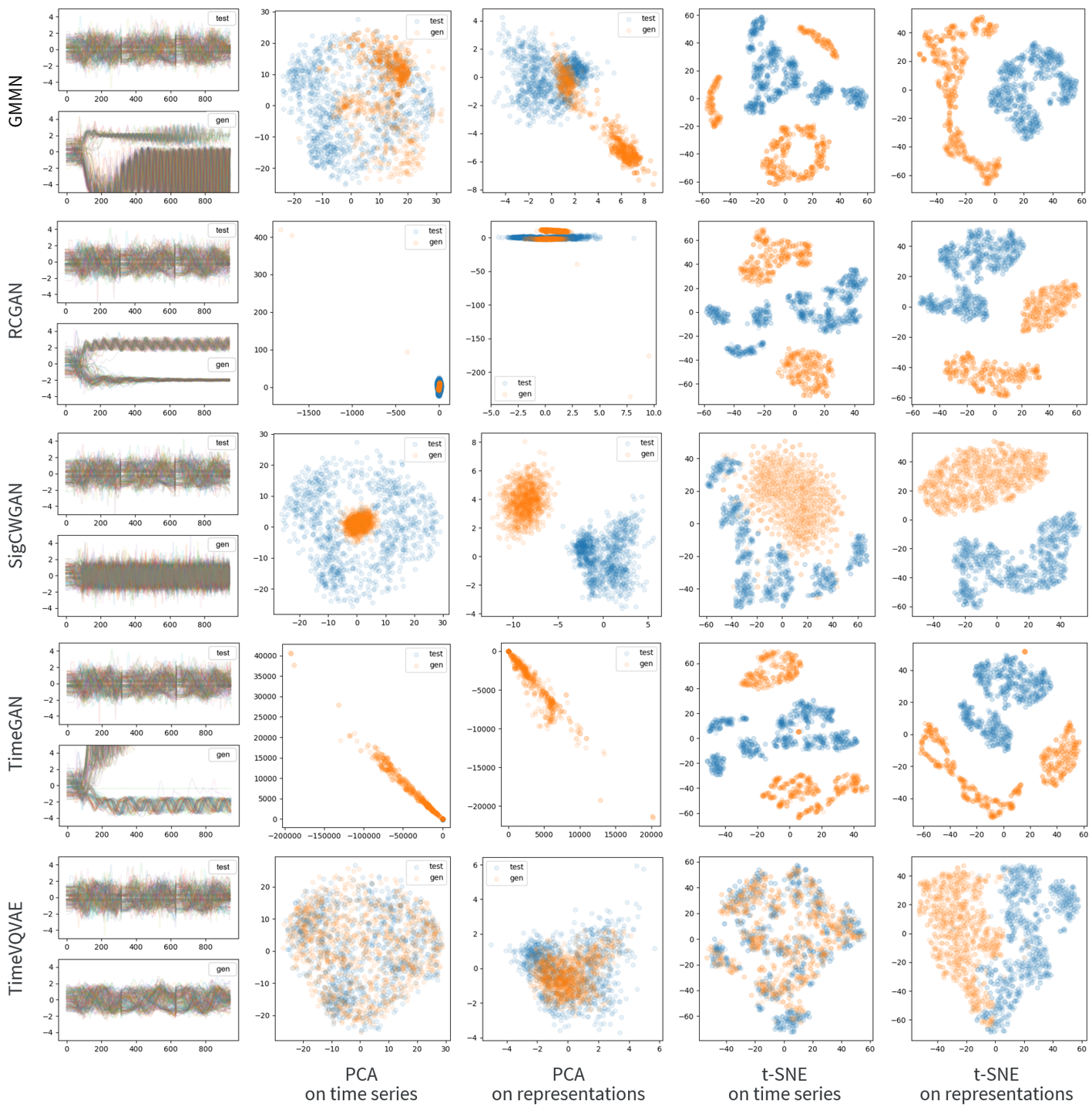}
\caption{PCA and t-SNE analyses. \textit{on time series} represents mapping of $x$. \textit{on representations} represents mapping of the representations of $x$. \textit{test} and \textit{gen} denote test and generated samples, respectively. The dataset of interest is UWaveGestureLibraryAll. The distribution differences for TimeVQVAE are partially due to train-test distribution difference which is unavoidable. FID scores are 737, 10146, 586, 726830513, and 4.7 for GMMN, RCGAN, SigCWGAN, TimeGAN, and TimeVQVAE, respectively.}
\label{fig:pca_tsne_analyses}
\end{figure}

The most common evaluation protocol in the TSG literature is PCA and t-SNE analyses on time series to visually see similarity between two distributions, one from real samples and the other from synthetic samples \citep{brophy2021generative, yoon2019time, li2022tts, zha2022time, desai2021timevae}. They, however, cannot be reported as a single score, which limits their usability. 
Also, because each element of time series does not carry semantically-meaningful information but time-step information, the principal axes found by PCA are not effective in capturing the realism of time series.
FID score, however, measures similarity between distributions of representations of time series. Since the representations capture high-level semantics of the input, the analyses on the representations can provide a better idea about the level of realism of generated samples.
Fig. \ref{fig:pca_tsne_analyses} presents the PCA and t-SNE analyses for TimeVQVAE and its competing methods. When visually inspecting the real and generated samples by TimeVQVAE, some difference is observable. The difference, however, is not clear on the PCA or t-SNE analyses on the time series but is clear on the representations. This shows the superiority of using FID score over the existing evaluation protocol.

\vfill

\end{document}